%% file: main.tex
\documentclass{article}

\usepackage{longcat_style}
\usepackage[export]{adjustbox}
\usepackage[utf8]{inputenc} %
\usepackage[T1]{fontenc}    %
\usepackage{newunicodechar}
\usepackage{hyperref}       %
\usepackage{xcolor}
\usepackage[normalem]{ulem} %
\hypersetup{
    colorlinks=true,      %
    linkcolor=blue,      %
    urlcolor=blue,       %
    citecolor=blue,      %
    linkbordercolor=blue, %
    urlbordercolor=blue,
    citebordercolor=blue,
    pdfborderstyle={/S/U/W 1}, %
}

\usepackage{float}
\usepackage{url}            %
\usepackage{booktabs}       %
\usepackage{amsfonts}       %
\usepackage{nicefrac}       %
\usepackage{microtype}      %
\usepackage{lipsum}		%
\usepackage{graphicx}
\usepackage{natbib}
\usepackage{doi}
\usepackage{amsmath}
\usepackage{amssymb} %
\usepackage{xspace}
\usepackage{enumitem}
\usepackage{multirow}
\usepackage{subcaption} 
\usepackage{makecell}
\usepackage{hyperref, cleveref}
\usepackage{pifont}
\usepackage[inkscapelatex=false]{svg}
\usepackage{caption}
\usepackage{wrapfig}
\usepackage{algorithm}
\usepackage{algorithmic}
\usepackage{threeparttable}
\usepackage{overpic}

\setlist[itemize]{leftmargin=*}
\setlist[enumerate]{leftmargin=*}
\setlist[description]{leftmargin=*}

\definecolor{midnightgreen}{rgb}{0.0, 0.29, 0.33}

\title{\textnormal{LongCat-Video-Avatar 1.5 Technical Report }}

\author{
    Meituan LongCat Team
}

\renewcommand{\headeright}{\raisebox{-0.2\height}{\includegraphics[height=1.8em]{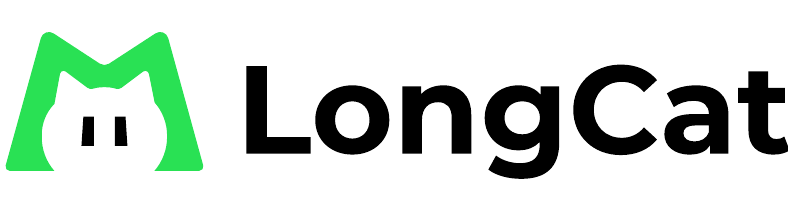}}}
\renewcommand{\shorttitle}{LongCat-Video-Avatar 1.5 Technical Report}

\usepackage{amsmath}
\begin{document}
\maketitle

\input{sec/0_abstract}

\clearpage
\tableofcontents
\clearpage

\input{sec/1_introduction}
\input{sec/2_dataset}
\input{sec/3_method}

\input{sec/4_training}
\input{sec/5_evaluation}

\input{sec/6_conclusion}
\input{sec/7_contributors}

\bibliographystyle{unsrtnat}
\bibliography{references}  

\clearpage
\appendix

\end{document}

%% file: sec/0_abstract.tex
\begin{abstract}

Despite advances in audio-driven video generation, achieving commercial-grade stability remains challenging. We present LongCat-Video-Avatar 1.5, an upgraded open-source framework prioritizing systematic engineering and production-readiness over architectural novelty.

By upgrading the audio encoder to Whisper Large and meticulously scaling our training recipes, v1.5 achieves accurate lip-synchronization, full-body temporal stability, and robust long-video generation with strict identity consistency. Through rigorous data curation and RLHF Training, the model readily generalizes to stylized domains such as anime and animals, and natively handles complex real-world conditions—such as multi-person interactions and object handling. Furthermore, addressing the practical demands of industrial deployment, we employ advanced step distillation to accelerate inference to an optimal 8 NFE, achieving a favorable trade-off between serving efficiency and visual fidelity.

The superiority of our approach is validated through extensive quantitative metrics and a rigorous human evaluation conducted on a comprehensive benchmark of over 500 diverse test cases. 
Results show that v1.5 achieves competitive or superior performance compared to leading closed-source systems (e.g., HeyGen, OmniHuman 1.5, Kling Avatar 2.0) across human-likeness ratings and expert-level quality assessments on our benchmark. With its open-source release, LongCat-Video-Avatar 1.5 narrows the gap between academic research prototypes and commercial-grade deployment.

\vspace{0.1cm}

\textbf{Page}: \href{https://meigen-ai.github.io/LongCat-Video-Avatar-1.5-Page}{https://meigen-ai.github.io/LongCat-Video-Avatar-1.5-Page} \\
\textbf{GitHub}: \href{https://github.com/meituan-longcat/LongCat-Video}{https://github.com/meituan-longcat/LongCat-Video}

\begin{figure}[htbp]
  \centering
  
  \begin{subfigure}[c]{0.48\textwidth}
    \centering
    \includegraphics[width=\linewidth]{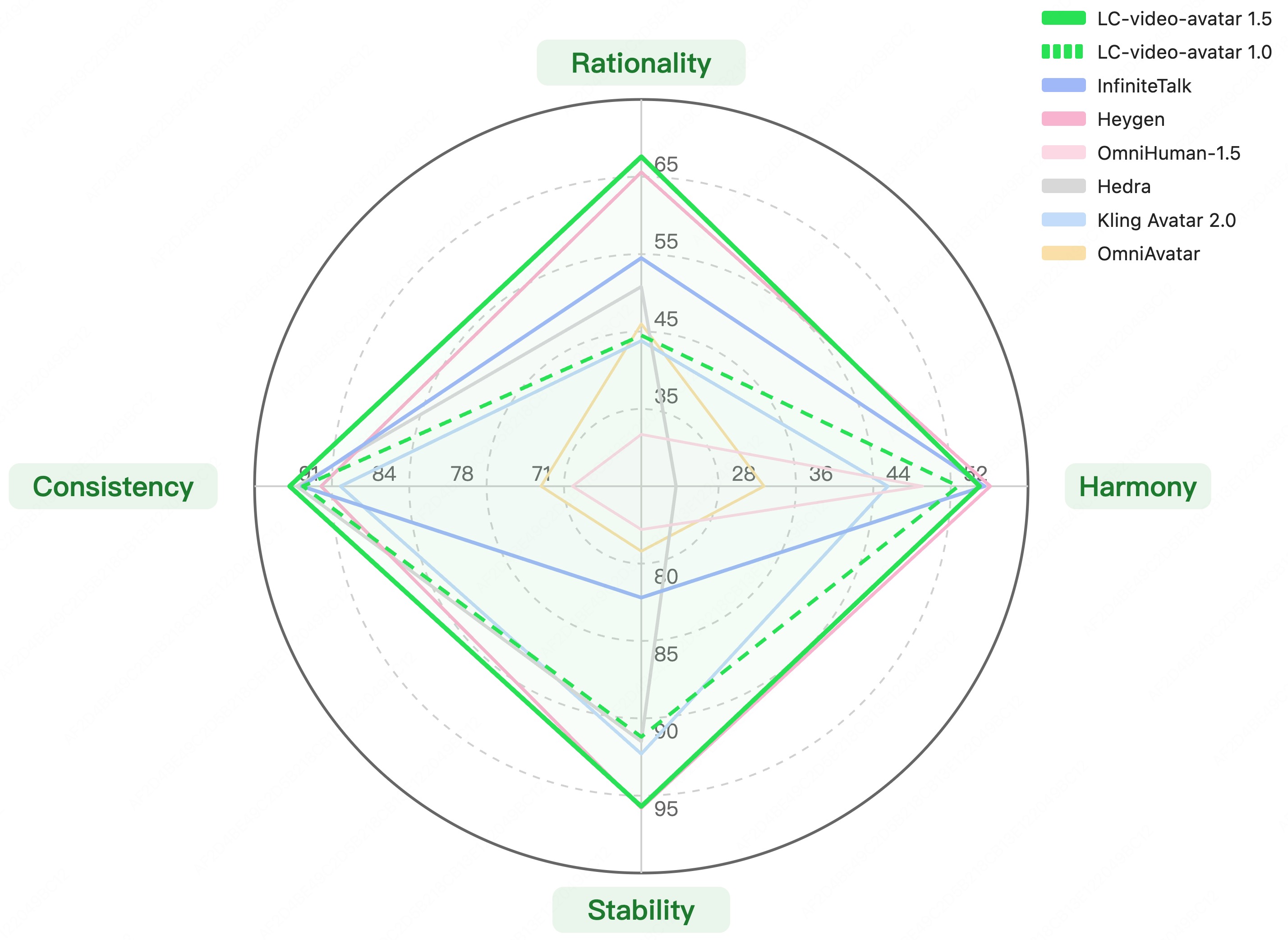}
    \caption{} 
    \label{fig:comp1}
  \end{subfigure}
  \hfill 
  \begin{subfigure}[c]{0.5\textwidth}
    \centering
    \includegraphics[width=\linewidth]{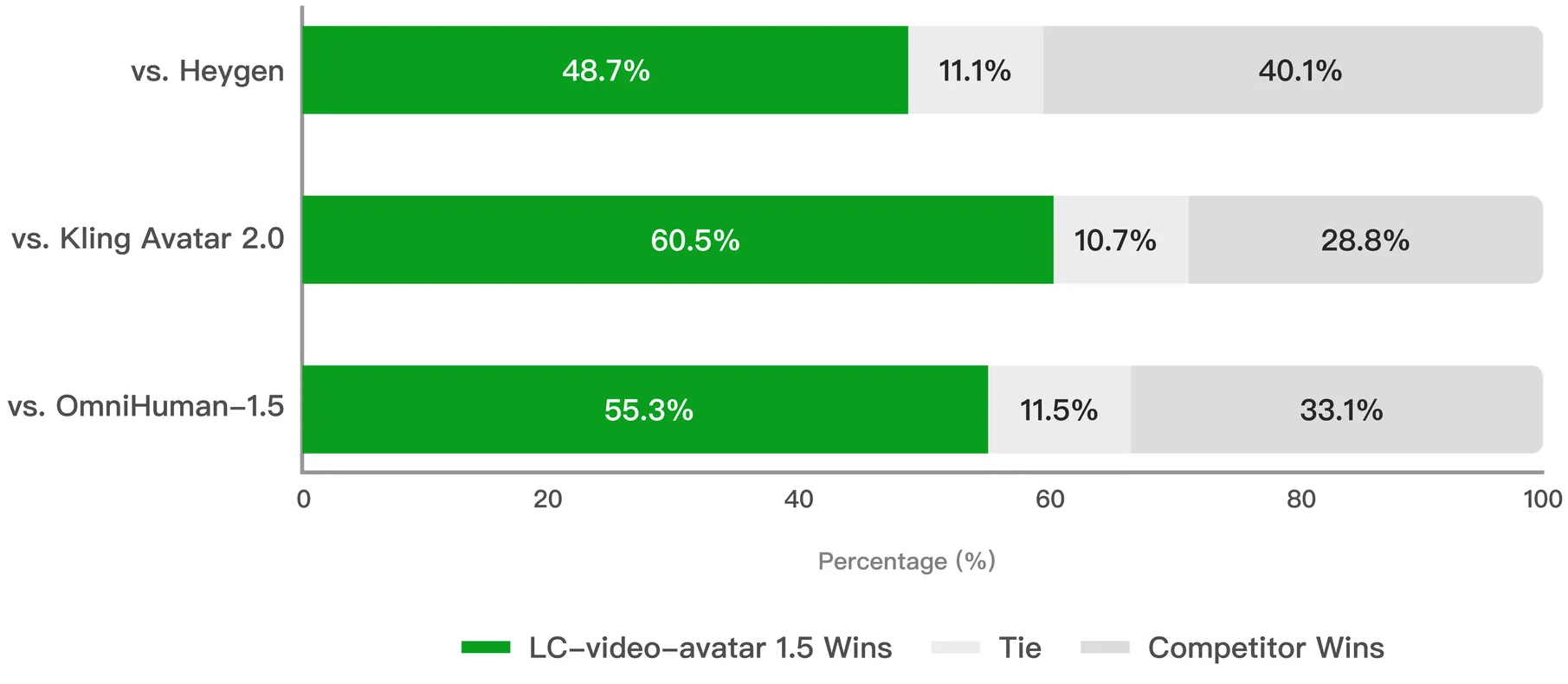}
    \caption{}
    \label{fig:comp2}
  \end{subfigure}
  
  \caption{Human evaluation. The overarching benchmark includes over 500 test samples with varying audio-visual complexities, scenarios, and languages. For the results in the two figures, only images containing a single talker are evaluated. (a) Expert-level objective quality evaluation across four dimensions~\cite{zhou2025evaltalker}, calculated as 100 $-$ Issue Rate, \textit{i.e.,} rationality, stability, harmony, and consistency. Issue rate means the percentage of samples rated as having the corresponding artifact by expert evaluators. (b) Human-likeness comparison with leading commercial models.}
  \label{fig:comp}
  
\end{figure}

\end{abstract}

%% file: sec/1_introduction.tex
\section{Introduction}

Audio-driven human animation aims to synthesize photorealistic avatar videos in which lip motion, facial expression, head pose, and body dynamics evolve coherently with a speech signal. As a core capability for digital humans, virtual communication, and embodied interactive systems, it has attracted increasing attention from both academia and industry. Recent progress in large-scale generative modeling, especially diffusion-based video generation \cite{wang2025fantasytalking,kong2025let,jiang2025omnihuman,team2025longcat}, has significantly improved visual fidelity, motion realism, and short-range temporal coherence, making audio-driven avatar generation a rapidly advancing frontier. This progress has catalyzed a surge of novel audio-driven generation methods \cite{gao2025wan,yang2025infinitetalk,chen2025hunyuanvideo,kong2025let,gan2025omniavatar, flowvqtalker}, including several recent efforts focused on real-time synthesis \cite{li2025joyavatar,li2025personalive,huang2025live,shen2025soulxflashtalktechnicalreport,zeng2026lpm}.

A substantial gap remains between research-quality demos and production-ready systems. In practice, commercial deployment requires far more than visually plausible short clips. A usable system must maintain stable identity over long durations, preserve full-body temporal consistency, synchronize lip movement precisely under diverse speaking styles, and remain robust in challenging real-world scenarios such as multi-person interactions, hand-object contact, stylized characters, and non-ideal source images. At the same time, the model must be efficient enough for cost-sensitive serving. These requirements expose a central tension in current audio-driven video generation: models that perform well on curated benchmarks may exhibit degraded robustness under long-horizon or open-domain conditions, while systems with strong real-world performance are typically proprietary and inaccessible to the broader community.

In this report, we present \textbf{LongCat-Video-Avatar 1.5} (LC-Video-Avatar 1.5), an upgraded open source framework designed to bridge this gap. To address the practical demands of commercial-grade digital human applications, this work focuses on generation stability and robustness in real-world scenarios. We demonstrate that a highly reliable, production-ready system can be effectively achieved through rigorous data curation, scaled model training, and comprehensive end-to-end optimization.

Specifically, to enhance the human-likeness of speech-driven animations, we upgrade our primary audio encoder to Whisper-large \cite{radford2022whisper}. This decision is driven by our empirical comparisons, which reveal that Whisper-large yields significantly smoother and more natural lip dynamics than the commonly used Wav2Vec2. When combined with our rigorous data and training pipelines, this architectural shift improves the model's ability to capture fine-grained speech dynamics, leading to markedly stronger lip synchronization and temporal smoothness over long videos. Moreover, these improvements generalize across diverse visual domains and complex scenarios beyond the training distribution. The model generalizes robustly across diverse visual domains (e.g., stylized anime characters and animals) and complex real-world situations (e.g., multiple people in frame and object interactions), all without requiring scenario-specific architectural branches.

Beyond foundational generation capabilities, we also emphasize the practical requirements of industrial deployment. To improve perceptual quality and align outputs with human preferences, we employ Group Relative Policy Optimization (GRPO) to significantly elevate the overall generation quality. However, diffusion-based video synthesis is typically constrained by high inference costs, which limit scalability in real serving environments. To address this bottleneck, we subsequently adopt advanced Distribution Matching Distillation (DMD) \cite{yin2024improved}, compressing the inference process to a highly efficient \textbf{8 NFEs}. This two-stage optimization pipeline—first enhancing quality via GRPO, then accelerating inference via DMD—achieves a favorable balance between visual excellence and serving cost, establishing LongCat-Video-Avatar 1.5 as a practically deployable open-source solution.

We validate the proposed framework through extensive quantitative evaluation and a rigorous human study on a benchmark  \cite{zhou2025evaltalker} of more than 500 diverse test cases. Across realism, naturalness, stability, and overall preference, LongCat-Video-Avatar 1.5 consistently outperforms strong baselines, including leading closed-source systems such as OmniHuman 1.5 \cite {jiang2025omnihuman} and HeyGen \cite{HeyGen}. These results suggest that, with sufficiently careful system-level optimization, open-source audio-driven avatar generation can move beyond research prototypes and begin to meet the demands of versatile commercial applications.

The main contributions of this report are summarized as follows:
\begin{itemize}
\item We introduce LongCat-Video-Avatar 1.5, a commercial grade, open-source framework for audio-driven video generation. Driven by rigorous data curation and scaled training recipes, our system achieves strong performance across multiple dimensions: precise lip synchronization, full-body temporal stability, strict identity consistency in long videos, and robust open-domain generalization to stylized characters and complex scenarios.
\item We achieve an optimal trade-off between generation quality and serving efficiency by implementing a step-distilled inference pipeline that requires only 8 NFEs. Additionally, we integrate GRPO to further elevate the generation quality.
\item Extensive evaluations—comprising both comprehensive automatic metrics and rigorous human studies on a large-scale benchmark—demonstrate that our efficient model consistently outperforms state-of-the-art closed-source alternatives in terms of naturalness and realism.
\end{itemize}

\begin{figure*}[t]
\centering
\includegraphics[width=0.8\textwidth]{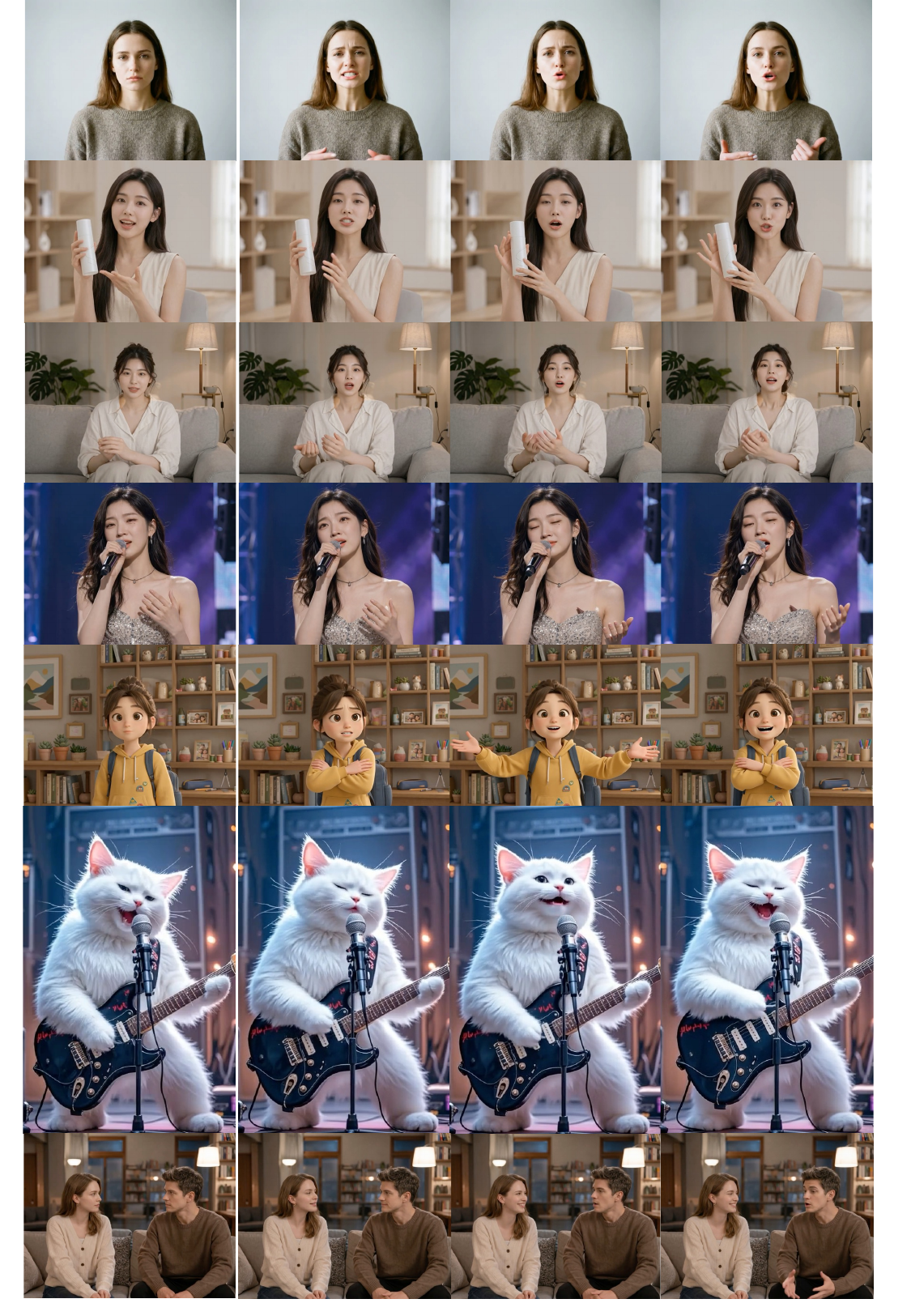} 
\vspace{-0.3cm}
\caption{Demonstration of generated video frames across various application scenarios, including broadcasting, acting, singing, e-commerce marketing, multi-person conversation, animation, and animal. The leftmost column shows the input, followed by the generated intermediate frames.}
\label{fig:scene_app}
\vspace{-1cm}
\end{figure*}

%% file: sec/2_dataset.tex
\section{Data}

\begin{figure*}[t]
\centering
\includegraphics[width=0.95\textwidth]{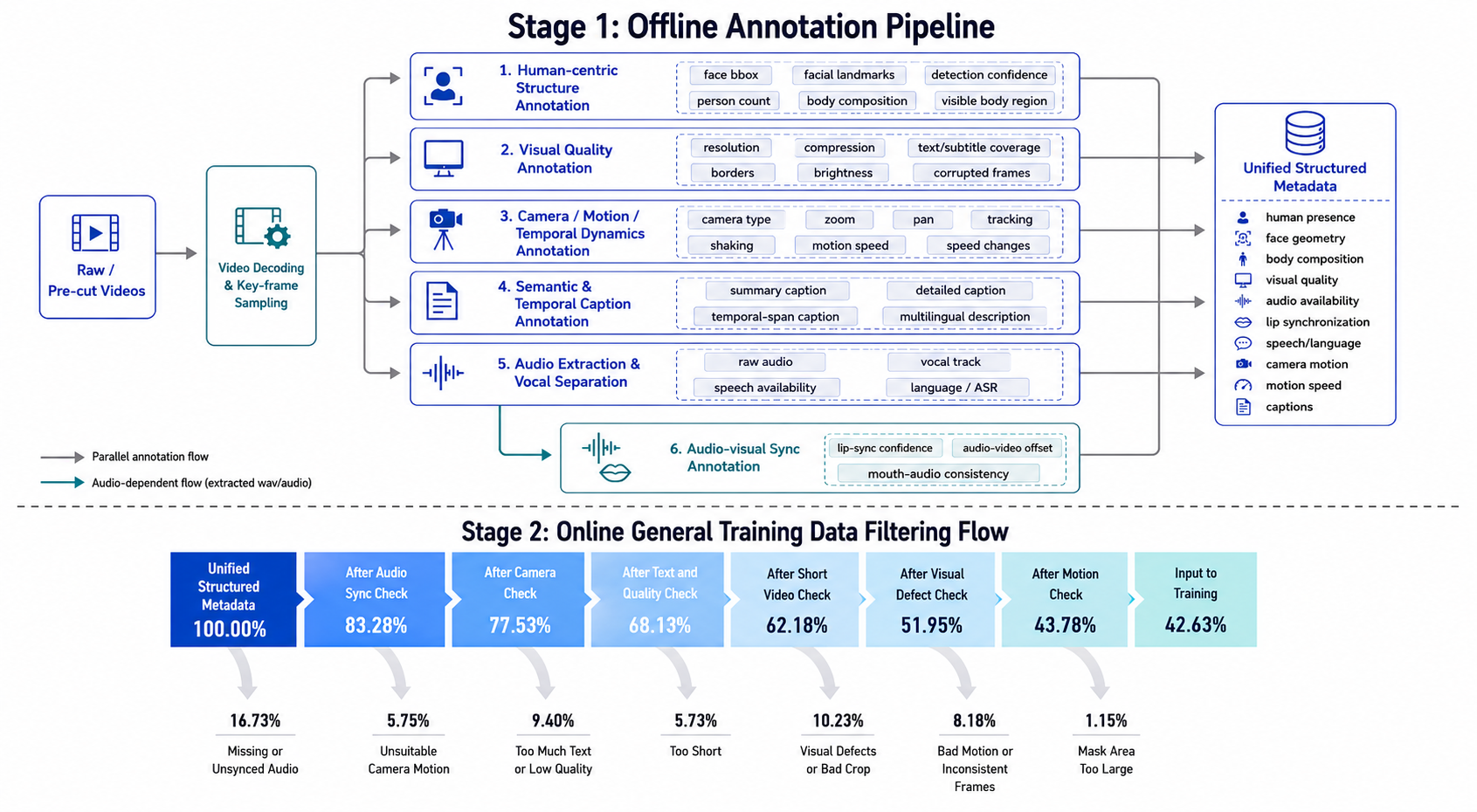} 
\vspace{-0.3cm}
\caption{Overview of the two stage data curation pipeline.}
\label{fig:general_data_flow}
\end{figure*}

\subsection{General Pipeline}

To enable LongCat-Video-Avatar 1.5 to generate stable and controllable single-person avatar videos, we build a multi-stage general data pipeline for large-scale single-person video data. This pipeline serves as the foundation of the overall data system and supports core model capabilities, including identity preservation, audio-driven lip motion, natural facial expressions, upper-body and full-body motion, hand-object interaction, camera-motion control, and style generalization.

\paragraph{Data Source Design.}
To cover the core capabilities required by LongCat-Video-Avatar 1.5, we organize raw videos according to their functional contribution to training rather than simply mixing them by source. \textbf{(1)} Close-up face videos are used to strengthen facial modeling, especially lip motion, expression details, and identity consistency. \textbf{(2)} Interview videos typically contain stable subjects, clear speech, and explicit talking states, providing reliable audio-visual correspondence for audio-driven training. \textbf{(3)} Acted performance videos contain richer camera language, pose variation, and scene dynamics, improving generalization to natural expressions, non-template motions, and complex scenes. \textbf{(4)} Interaction videos cover object holding, pointing, manipulation, and conversational gestures, improving the naturalness of hand motion and human-object interaction. \textbf{(5)} Music videos contain singing, rhythmic motion, stage performance, and high-intensity expressions, complementing rhythm-driven and performance-oriented avatar scenarios. \textbf{(6)} Animation and stylized videos further expand the model's ability to generalize to non-photorealistic appearances and stylized character forms.

Although these data sources are complementary, their distribution gaps also introduce substantial noise. Videos from different sources vary significantly in face scale, body composition, camera motion, visual quality, audio condition, language distribution, and caption granularity. Directly mixing them for training may introduce non-human subjects, multi-person ambiguity, audio-visual misalignment, low-quality frames, border or subtitle artifacts, abnormal speed changes, and mismatches between captions and sampled local clips. The primary objective of the general pipeline is to transform heterogeneous videos into structurally consistent, quality-controlled, and semantically aligned training samples, rather than simply aggregating raw data at scale.

\paragraph{Unified Annotation Schema.}
To incorporate heterogeneous videos into a unified training framework, we design a unified annotation schema that converts implicit video attributes into structured metadata. The schema covers human presence, face geometry, body composition, visual quality, audio availability, lip synchronization, speech and language, camera motion, motion speed, and semantic captions. With this schema, videos from different sources are mapped into a comparable and reusable data representation space, allowing subsequent training stages to select data based on content, quality, and conditioning attributes rather than coarse source-level rules.

$\bullet$ \textbf{Offline Annotation.}
The goal of offline annotation is to establish a unified data understanding layer. This stage processes full videos or pre-cut clips and uses visual, audio, and multimodal models to extract relatively stable content and quality attributes. Since these attributes are independent of the specific training clip sampled online, they can be precomputed and reused across different training configurations. As illustrated in the first part of Fig.~\ref{fig:general_data_flow}, most annotation modules are executed in parallel to construct reusable metadata, while audio-visual synchronization is performed after audio extraction and vocal separation.

For human-centric structure annotation, we annotate face location, facial landmarks, detection confidence, person count, visible body region, and body composition. Close-up face data relies on these annotations for face localization and head-pose filtering, ensuring that the mouth region is clearly visible. Upper-body and full-body data use body composition annotations to distinguish head, half-body, and full-body samples, preventing different human scales from being mixed without control. This allows different training stages to use samples that match their target composition, improving the stability of facial detail modeling, identity consistency, and body-motion learning.

For audio and lip-sync annotation, we extract raw audio, separate vocal tracks, and estimate audio-visual synchronization for talking videos. Audio annotation verifies whether a sample contains usable speech conditions, while lip-sync annotation measures the temporal consistency between vocal audio and mouth movement. Samples with large audio-visual offsets or low synchronization confidence are removed, since they corrupt the supervision between audio and lip motion and directly degrade lip-sync generation quality.

For visual quality annotation, we estimate perceptual video quality and describe common artifacts such as text coverage, borders, black borders, abnormal brightness, and pixel-level degradation. This stage identifies low-resolution, heavily compressed, subtitle-heavy, black-border, white-flash, transition, and locally corrupted samples, providing a unified basis for later quality filtering. The purpose is to prevent the model from learning blurred textures, compression artifacts, or abnormal boundary patterns.

For camera, motion, and temporal dynamics annotation, we identify camera type, camera motion, and motion speed. Music and acted performance videos often contain zooming, panning, tracking, shaking, rhythmic motion, or editing-induced speed changes, while interview and close-up face videos are usually more static. Explicitly annotating these attributes enables later training stages to select static-camera or natural-speed samples when needed, and also allows camera-motion information to be injected into textual conditions as a controllable signal.

For semantic and temporal caption annotation, we generate multilingual and multi-granularity video descriptions, including detailed captions, summary captions, and temporal-span captions. For short videos, global captions usually describe the main content sufficiently. For acted performance and long-form videos, however, a global caption may not correspond to the local clip sampled during training. We therefore introduce temporal-span captions so that each sampled clip can be paired with a more accurate local description, reducing text-video misalignment.

$\bullet$ \textbf{Online Clip-Level Validation and Condition Construction.}
The online stage ensures that each sampled training clip satisfies quality requirements and converts offline annotations into task-specific training conditions. As illustrated in the second stage of Fig.~\ref{fig:general_data_flow}, candidate metadata are progressively filtered by audio synchronization, camera suitability, text and visual quality, duration, visual defects, motion consistency, and mask-area constraints before being used as training inputs. This staged design makes the filtering process interpretable and allows us to identify the dominant sources of data removal at each step. Offline annotations usually describe a full video or a pre-cut clip, whereas training uses a local temporal window sampled from the video. Even if the video is globally valid, the sampled window may still contain transitions, black frames, white flashes, under-exposure, over-exposure, residual borders, sudden frame jumps, or abnormal motion. The online stage therefore serves as the final clip-level quality control layer.

For task-specific sample selection, different training objectives use different combinations of offline annotations. Close-up face training emphasizes face visibility, head pose, and lip synchronization. Upper-body and full-body training focuses more on body composition, hand visibility, and camera stability. Complex-scene and acted performance data are more based on local caption alignment, visual quality, and temporal continuity. Music and interaction data emphasize rhythmic motion, expressive motion, and natural hand-object interaction. In this way, the same general data pool can provide multiple task-oriented subsets for different training stages of LongCat-Video-Avatar 1.5.

For clip-level quality validation, we perform a second-stage check on the actual sampled clip, including duration, sampling frame rate, resolution, brightness distribution, black/white pixel ratio, border artifacts, frame jumps, and motion intensity. This mechanism directly validates the temporal window that enters training. It improves filtering precision, avoids discarding an entire long video due to a few corrupted segments, and balances data quality with data utilization.

For condition construction, part of the structured annotations is converted into textual conditions. The motion of the camera, the size of the shot, the type of lens and the visual style can be combined with the original caption, so the final text condition describes not only the visual content but also the shooting style and the camera behavior. This makes implicit controllable factors explicit, enabling LongCat-Video-Avatar 1.5 to learn the relationship among semantic content, human motion, and camera language.

Overall, the general pipeline transforms complex heterogeneous videos into unified, interpretable, filterable, and reusable training data. Through data source design, offline annotation, online clip-level validation, and condition construction, it establishes the general-purpose data foundation for LongCat-Video-Avatar 1.5. This foundation pipeline is essential for stable avatar generation across identity preservation, lip synchronization, body motion, camera control, and style generalization. However, we observe that existing avatar generation models, including MultiTalk~\cite{kong2025let}, OmniHuman 1.5~\cite{jiang2025omnihuman} and LongCat-Video-Avatar 1.0~\cite{meituanlongcatteam2025longcatvideoavatartechnicalreport}, still show noticeable limitations in several challenging scenarios, especially multi-person interaction, silent non-speaking motion, and emotional expression. To further improve generation quality in these under-addressed settings, we design three specialized data pipelines for multi-person, silent, and emotion-specific data on top of the general data framework.

\subsection{Multi-Person Data}
We develop a multi-person data curation pipeline that transforms raw videos into structured audio-visual supervision for multi-speaker modeling. The pipeline first applies ByteTrack-based person tracking~\cite{zhang2022bytetrack} to extract person-level spatio-temporal trajectories. This track-level filtering separates dynamic human subjects from static human-like artifacts, such as portraits or posters, and enables videos to be categorized into single-person and multi-person subsets.

For multi-person videos, we employ an active speaker detection (ASD) pipeline built upon established audio-visual ASD methods, including TalkNet and UniTalk~\cite{tao2021talknet,nguyen2025unitalk}. In our optimized implementation, YOLOv6~\cite{li2022yolov6} is used as an efficient real-time detection backbone. The ASD stage predicts the speaking intervals for each visible subject together with the corresponding face bounding-box trajectories.

These spatio-temporal annotations provide track-level speaker activity labels, specifying the temporal speech regions associated with each visible face trajectory. We further use these labels to exclude intervals with concurrent speaker activity and retain non-overlapping single-speaker segments, thereby reducing speaker ambiguity in the training data.

\subsection{Silent Data}
We develop a silent-video curation pipeline to collect non-speaking human videos for silent avatar generation. This data is complementary to audio-driven talking data: instead of learning the correspondence between speech and lip motion, it teaches the model to preserve natural facial stillness and non-verbal motion when no speech is present. Such samples are important for suppressing unintended mouth movement and for modeling gaze shifts, head motion, posture changes, gestures, and hand-object actions in silent scenarios.

The pipeline first decomposes long videos into short temporal clips, since a single video may contain both speaking and non-speaking intervals. Each clip is then analyzed independently to determine whether the visible subject is speaking. This clip-level design avoids assigning a coarse video-level silent label to content with mixed temporal states.

To improve the reliability of silent-state recognition, we use a two-stage multimodal verification strategy. In the first stage, we employ \texttt{Qwen3-Omni}~\cite{xu2025qwen3} to perform an initial assessment of whether the visible subject is silent. In the second stage, we use \texttt{Qwen3-VL}~\cite{Qwen3-VL} to independently re-evaluate the same clip. A clip is retained as silent only when both models agree that the subject is not speaking. This conservative agreement rule reduces false positives caused by brief speech, ambiguous mouth motion, singing-like articulation, off-screen speech, or unstable predictions.

After clip-level verification, we aggregate the decisions across the full video. Videos whose sampled temporal clips are consistently classified as non-speaking are retained as silent data, while videos containing any detected speaking interval are excluded from the silent subset. This strict aggregation strategy ensures that the resulting data provides clean supervision for non-verbal avatar motion.

The curated silent subset is used to strengthen silent and prompt-driven generation. By explicitly separating non-speaking videos from audio-driven talking videos, LongCat-Video-Avatar 1.5 learns to maintain inactive mouth motion when speech is absent, while still generating natural facial micro-motion, head movement, body motion, and interaction behaviors.

\begin{figure*}[t]
\centering
\includegraphics[width=0.9\textwidth]{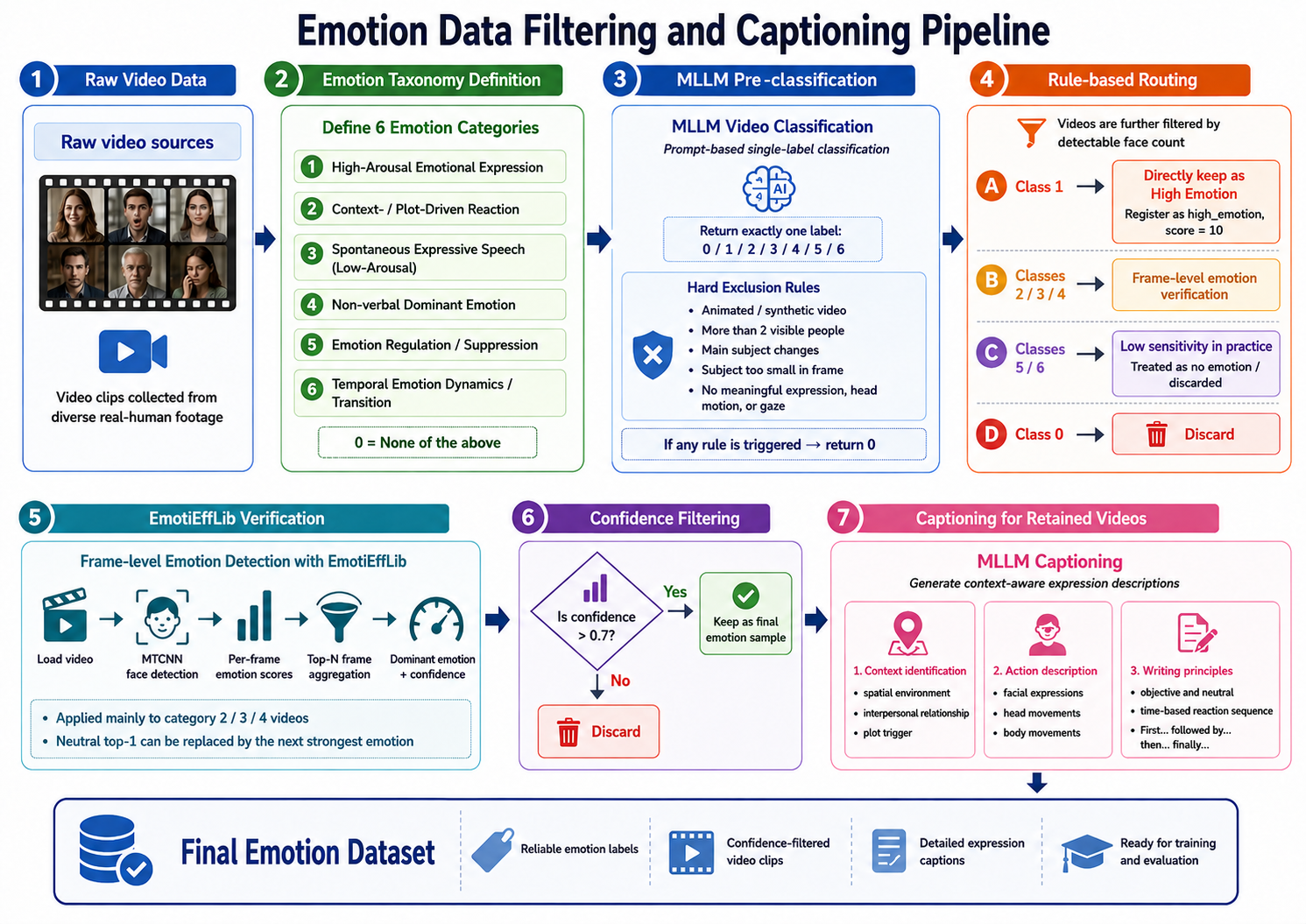} 
\caption{Overview of the Emotion Data Filtering and Captioning Pipeline. } 
\label{fig:emotion_data}
\end{figure*}

\subsection{Emotion Data}
To ensure that LongCat-Video-Avatar 1.5 can generate expressive and nuanced character motions, we develop a multi-stage curation pipeline specifically for emotional content as shown in Fig.~\ref{fig:emotion_data}. Unlike standard datasets that often focus on static facial expressions, our approach prioritizes the temporal evolution of emotion and its relationship with speech and context.  \textbf{Emotion Taxonomy and Initial Tagging.}  We first define a taxonomy of six distinct emotional categories to capture the breadth of human expression: 
\begin{enumerate}     
    \item {High-Arousal Emotional Expression:} High-intensity, large-amplitude states where the emotion type is clear and dominant (e.g., shouting, intense laughter).     
    \item {Context-/Plot-Driven Reaction:} Reactions triggered by external stimuli or dialogue, often characterized by a "reaction-before-expression" sequence (e.g., pauses, gaze shifts).     
    \item {Spontaneous Expressive Speech (Low-Arousal):} Natural, subtle emotional leakage during daily conversation, conveyed through continuous micro-expressions.     
    \item {Non-verbal Dominant Emotion:} Information primarily conveyed through facial movement, gaze, or posture rather than speech.     
    \item {Emotion Regulation/Suppression:} Instances where a character attempts to mask an underlying emotion, resulting in brief "micro-expression" flashes.     
    \item {Temporal Emotion Dynamics:} Videos where the emotion evolves through clear stages or turning points (e.g., transition from neutral to frustrated). 
\end{enumerate}  

We utilize \texttt{Qwen3-Omni}~\cite{xu2025qwen3} to perform initial classification. To ensure production quality, we implement a set of \textbf{Hard Exclusion Rules}. Any video containing synthetic content, more than two subjects, identity switches, or subjects occupying a small portion of the frame is assigned a null label (0).  For valid clips, the model assigns a category based on a priority hierarchy: $6 > 5 > 4 > 2 > 1 > 3$.  

\textbf{Refined Filtering with EmotiEffLib.}  Empirical observation reveals that while the LLM is highly effective at identifying high-arousal states (Category 1), it exhibits lower sensitivity toward Categories 5 and 6, often resulting in noisy samples. Furthermore, Categories 2, 3, and 4 contain a mix of expressive and near-neutral samples.  

To address this, we employ the \texttt{EmotiEffLib}~\cite{savchenko2023facial} framework for frame-level emotion recognition. Our filtering logic is as follows: 
\begin{itemize}     
    \item {Scoring:} We compute an emotion score by averaging the confidence of the top-$N$ frames ($N=10$ or $20$) for each emotion class.     
    \item {Neutral Bias Correction:} If "Neutral" is the top-1 predicted class, we automatically register the second-highest emotion class and its corresponding score to ensure we capture the underlying expressive signal.     
    \item {Confidence Thresholding:} We only retain videos where the final dominant emotion class achieves a confidence score of $s > 0.7$.  
\end{itemize}  

Videos originally tagged as Category 1 by the LLM are prioritized, while those in Categories 5 and 6 that fail to show significant emotional peaks in \texttt{EmotiEffLib} are re-classified as non-emotional data.  

\textbf{Context-Aware Annotation.}  The final filtered subset is processed through a specialized captioning pipeline to generate high-granularity descriptions. Unlike standard captions, our prompts require \texttt{Qwen3-Omni} to establish three levels of context: \textit{Spatial Environment}, \textit{Interpersonal Relationships}, and \textit{Plot Progression}.  

The resulting descriptions follow a principle of \textbf{Objective Neutrality}, focusing on physical manifestations rather than subjective interpretations. Annotations detail the chronological evolution of movement across three dimensions: 
\begin{itemize}     
    \item {Facial Expressions:} Forehead wrinkles, eyebrow position, gaze direction, and blink rates.     
    \item {Head Movements:} Displacement, tilt, rotation, and rhythmic swaying.     
    \item {Body Movements:} Posture shifts (e.g., leaning forward/backward), shoulder shrugging, and hand gestures. 
\end{itemize}  

This structured data ensures the model learns the causal relationship between a character's environment, their internal state, and their physical response.

%% file: sec/3_method.tex
\section{Method}
\subsection{Architecture}

\begin{figure*}[t]
  \centering
  \includegraphics[width=0.99\linewidth]{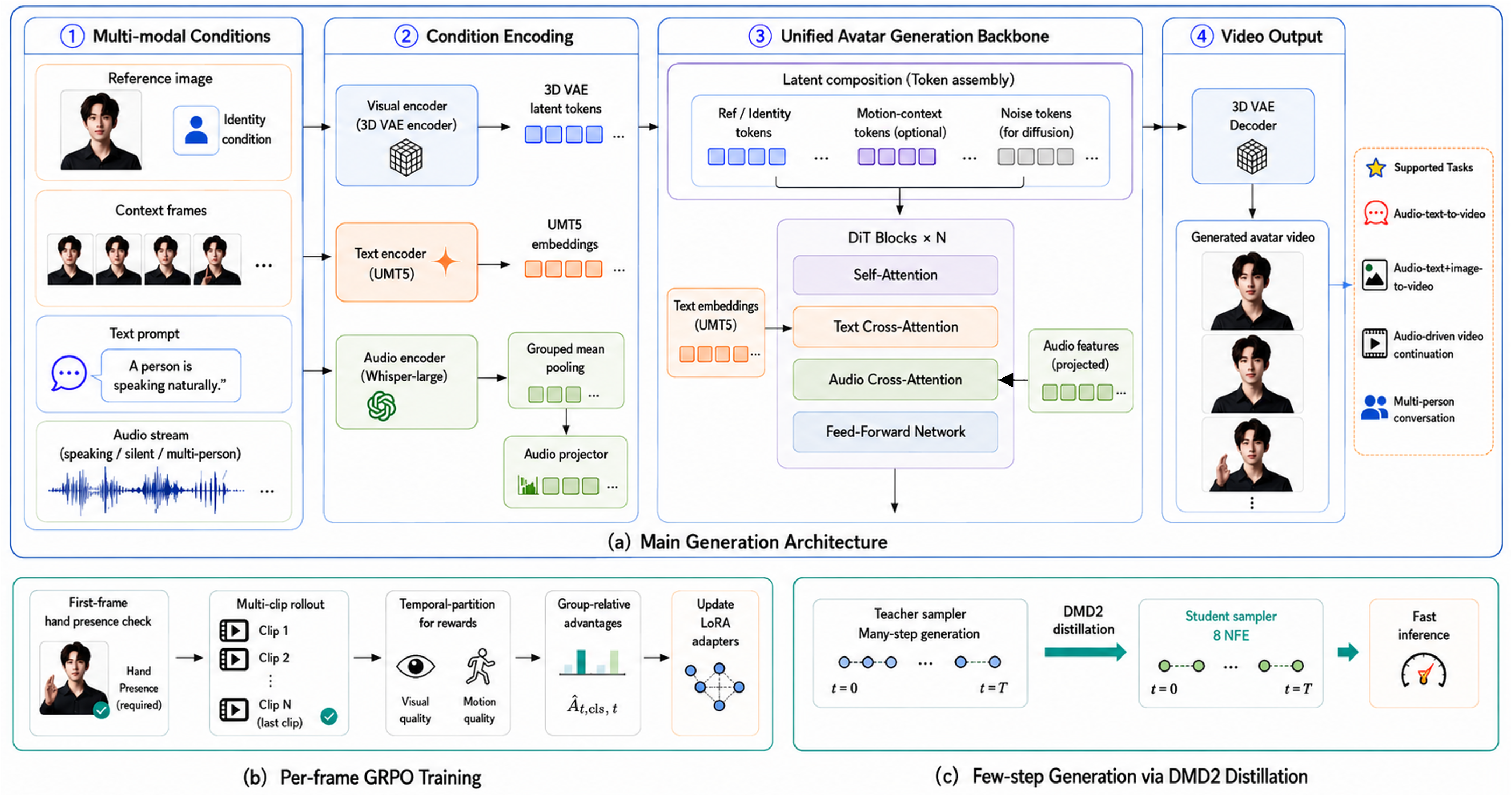}
  \caption{The overall pipeline of LongCat-Video-Avatar 1.5.}
  \label{fig:pipeline}
\end{figure*}

In this work, we inherit the unified DiT-based video diffusion architecture from LongCat-Video-Avatar 1.0 \cite{meituanlongcatteam2025longcatvideoavatartechnicalreport}. The model is built upon a 3D Variational Autoencoder (VAE), and each Diffusion Transformer (DiT) block comprises 3D self-attention, text cross-attention, and a Feed-Forward Network (FFN). Text embeddings are encoded using a UMT5 encoder, while 3D Rotary Position Embeddings (RoPE) are applied to the visual tokens to encode spatiotemporal positional information. The overall network architecture of the proposed method is shown in Fig.\ref{fig:pipeline}.

Our unified architecture supports multiple audio-driven human animation tasks with different input configurations. The network accepts three types of latent sequences as input: a reference latent, motion latents, and noise latents. For text-to-video generation, only noise latents are provided. For text-image-to-video generation, the reference latent is temporally concatenated with the noise latents. For video continuation, the context latents are temporally concatenated with the noise latents and fed into the model as additional conditioning signals.

To enable audio-driven generation within this unified video foundation model, we modify each DiT block by inserting an additional audio cross-attention layer after the text cross-attention module. This allows audio cues to be seamlessly integrated into the visual generation process. To prevent training instability and ensure the model effectively aligns audio signals with corresponding mouth movements without catastrophic forgetting of pre-trained visual priors, we retain the Adaptive Layer Normalization (adaLN) module before each audio cross-attention layer. This module functions as a gating mechanism that progressively incorporates audio control, thereby stabilizing optimization and facilitating the learning of accurate audio-to-lip motion mappings.

\subsection{Audio Feature Extraction}

In this version, we significantly upgrade our audio encoder from the Wav2Vec2 \cite{baevski2020wav2vec} model used in v1.0 to Whisper-large \cite{radford2022whisper}. Compared to the 94M-parameter Wav2Vec2, Whisper-large features 1.5B parameters and is pre-trained on 680,000 hours of multilingual speech data. This architectural enhancement yields substantially richer acoustic representations, superior phoneme level expressiveness, and stronger multilingual robustness, as it operates directly on Mel spectrograms extracted from raw audio waveforms.

To process audio streams exceeding Whisper's 30-second context limit, we adopt a sliding window strategy. The input spectrogram is partitioned along the time dimension and forwarded through the Whisper encoder, which yields 33 hidden states (the embedding layer plus 32 transformer layers) at an internal frame rate of 50 Hz. To compress this high-dimensional multi-layer output into a compact representation, we follow \cite{chen2025humo} and apply a grouped mean pooling strategy. Specifically, all 33 hidden states are divided into four groups of 8 layers each, plus one singleton layer. Each group is reduced via mean-pooling to form a 5 channel feature representation. Subsequently, these 5 channel features are temporally resampled via linear interpolation from 50 Hz to the target video frame rate of 25 FPS, yielding an audio embedding of shape $(T, 5, 1280)$, where $T$ denotes the number of video frames and 1280 is the hidden dimension. Finally, because the video VAE applies a $4\times$ temporal downsampling when converting pixel-space videos into the latent space, a corresponding temporal compression is required for the audio embeddings. We employ an audio projector that aggregates neighboring context within a temporal window and downsamples the 25 FPS audio features to match the latent sequence length. This ensures strict temporal alignment between the audio cues and visual latents prior to their injection into the audio cross-attention layers.

\begin{figure*}[t]
  \centering
  \includegraphics[width=0.9\linewidth]{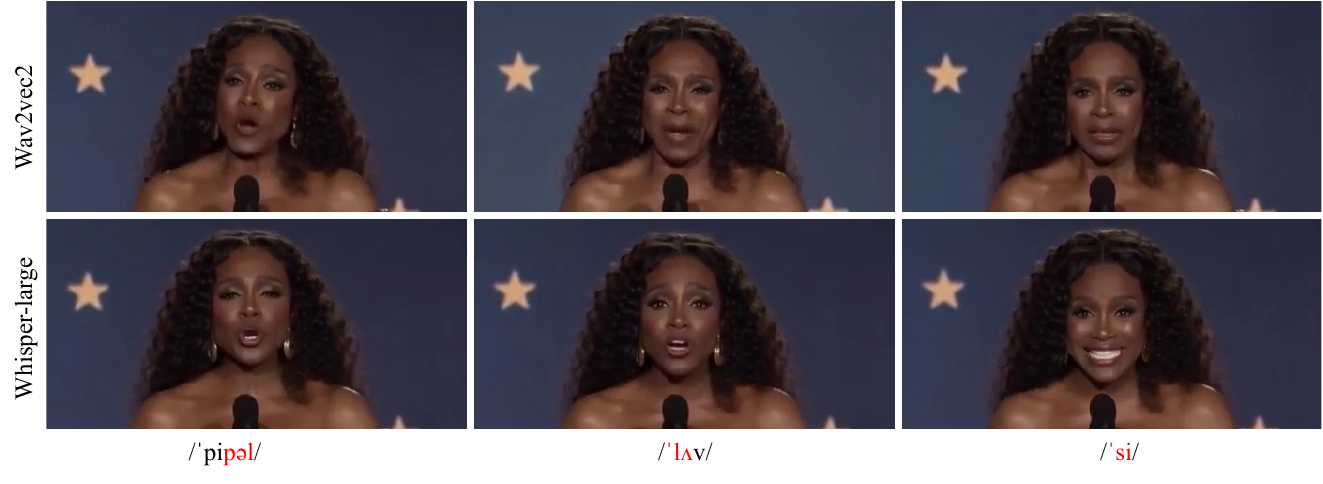}
  \caption{The lip synchronization comparison between Wav2vec2 and Whisper-large.}
  \label{fig:audio_encoder}
\end{figure*}

As illustrated in Fig.~\ref{fig:audio_encoder}, our comparison between Wav2Vec and Whisper-large clearly demonstrates that Whisper-large not only achieves highly accurate and fine grained audio-lip synchronization, but also produces much more natural and fluid mouth movements.

\subsection{Group-Relative Per-Frame Policy Optimization}
\label{sec:grpo}

Our training framework largely follows the multi-reward GRPO formulation of LongCat-Video~\cite{team2025longcat}. 
Our main extension is to move from video-level reward modeling to \emph{per-frame} reward modeling. In LongCat-Video, each reward model $R_k$ produces a video-level reward and the corresponding relative advantage is computed at the sample level. Here, we instead decompose each reward along temporal partitions. Let $r_{k,j}^i$ denote the reward of the $j$-th temporal partition of sample $i$ under reward model $R_k$. Following the same group-relative normalization strategy as LongCat-Video, we define
\begin{equation}
\hat A_{k,j}^i
=
\frac{r_{k,j}^i-\mu_{k,j}}{\sigma_{k,j}^{\max}},
\end{equation}
where $\mu_{k,j}$ is the group mean and $\sigma_{k,j}^{\max}$ is the maximum group standard deviation for reward $R_k$ at temporal partition $j$, following the stabilized normalization used in LongCat-Video.

Consistent with the multi-reward training strategy in LongCat-Video, the effective relative advantage is the weighted sum of the individual relative advantages:
\begin{equation}
\hat A_{\mathrm{total},j}^i
=
\sum_k w_k \hat A_{k,j}^i.
\end{equation}
Therefore, our method preserves the same multi-reward aggregation form as LongCat-Video, while extending the advantage from a video-level scalar to a temporally structured signal.
The resulting per-frame relative advantage is then used for diffusion policy optimization on stored denoising transitions. Compared with the original video-level formulation, this extension enables finer-grained credit assignment and allows the optimization to focus on temporally localized artifacts such as local motion inconsistency, hand deformation, and short-range structural collapse.

\textbf{First-frame hand-presence check.} For image-to-video, and video-continuation tasks, we further introduce a task-aware first-frame hand-presence check. Since hand quality can only be meaningfully supervised when the conditioning frame contains visible hands, we prioritize such samples during preference optimization, thereby increasing the proportion of hand-relevant training examples and helping alleviate hand distortion in conditioned human video generation.

\textbf{Multi-clip rollout.} To better support long-horizon video-continuation generation, we additionally adopt a multi-clip rollout strategy. Multiple clips are generated sequentially, where earlier clips are used to build temporal context and only the final clip participates in GRPO optimization. In this way, our method preserves the overall LongCat-Video training paradigm while extending it to per-frame reward-based credit assignment and longer temporal continuation.

\subsection{Few-step Generation}
\label{sec:fsg}

Inspired by the Distribution Matching Distillation 2 (DMD2) \cite{yin2024improved} framework, we distill multi-step diffusion models into efficient few-step generators. DMD2 aligns the generator’s distribution with the teacher's by minimizing the reverse Kullback-Leibler (KL) divergence. However, the standard implementation requires substantial GPU memory, as it requires maintaining three separate, homogeneous models in VRAM simultaneously: the generator, the fake score function, and the real score function. To overcome this VRAM bottleneck, we propose a parameter-efficient architecture that utilizes a single base Diffusion Transformer (DiT) backbone equipped with multiple LoRA adapters. Specifically, we employ a shared backbone and differentiate its functional roles by dynamically mounting either a Generator LoRA or a Fake Score LoRA. This design enables seamless switching between few-step sampling and score estimation, while the original base DiT provides the real-score guidance. To balance inference speed and generation quality, we distill the model to 8 denoising steps. During this process, both the real and fake score functions retain the time scheduler from the preceding training stages to ensure a consistent noise scale. Furthermore, to mitigate over-saturation typically observed during distillation, we slightly reduce the Classifier-Free Guidance (CFG) scales for both text and audio to 4.0. Our approach significantly reduces the hardware footprint while preserving the high-fidelity distribution matching performance of the original DMD2.

\begin{figure*}[t]
  \centering
  \includegraphics[width=0.99\linewidth]{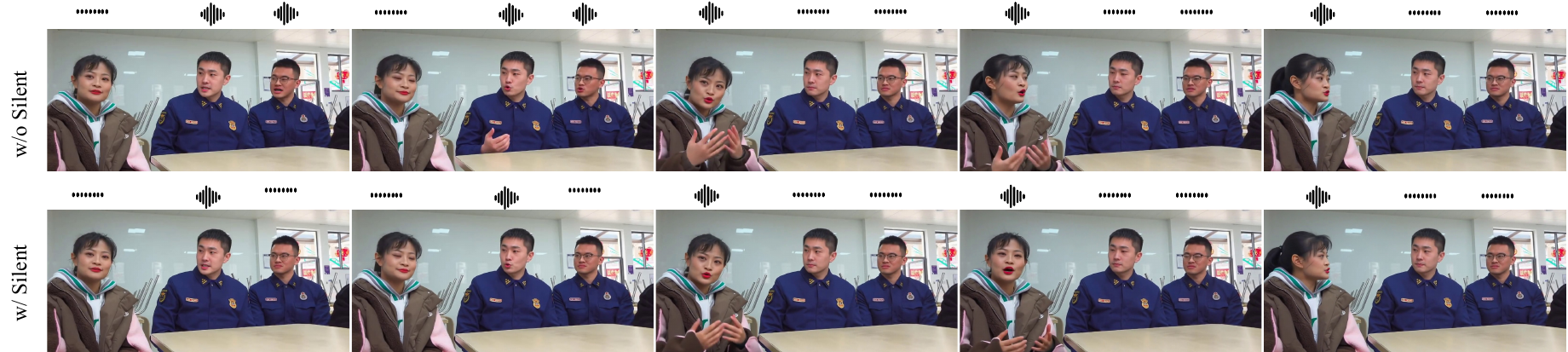}
  \caption{Visual illustration of the background character driving strategy. (a) w/o Silent Condition. (b) w/ Silent Condition.}
  \label{fig:multiti}
\end{figure*}

\begin{table}[t]
\centering
\small
\caption{Outline of the progressive training stages.}
\label{tab:training_stages}
\begin{tabular}{lcccc}
\toprule
\textbf{Training Tasks} 
    & \textbf{Size Bucket} 
    & \textbf{Batch Size} 
    & \textbf{Learning Rate} 
    & \textbf{Iterations} \\
\midrule
AT2V + AI2V + VC 
    & $256\text{p} \times 93$ frames 
    & 64
    & $2 \times 10^{-5}$ 
    & 130k \\
AT2V + AI2V + VC 
    & $480\text{p} \times 93$ frames 
    & 32
    & $2 \times 10^{-5}$ 
    & 45k \\
AT2V + AI2V + VC + Ref
    & $480\text{p} \times 93$ frames 
    & 32
    & $2 \times 10^{-5}$ 
    & 28k \\
AT2V + AI2V + VC + Ref
    & $480\text{p} + 720\text{p} \times 93$ frames 
    & 32
    & $2 \times 10^{-5}$ 
    & 6k \\
AT2V + AI2V + VC + Ref + MultiTalk
    & $480\text{p} + 720\text{p} \times 93$ frames 
    & 32
    & $2 \times 10^{-5}$ 
    & 2k \\
\bottomrule
\end{tabular}
\end{table}

\subsection{Multi-Person Conversation}

For two-person conversational video generation, we follow the training strategy of MultiTalk \cite{kong2025let} and adopt the L-RoPE mechanism to explicitly associate each speaker region with its corresponding audio condition. Meanwhile, reference attention maps are used to establish region-level correspondences between visual character regions and audio signals.

When multiple individuals appear in the reference image, we designate two of them as target speakers and treat the remaining individuals as background. However, this setting introduces an attribution ambiguity: due to the high visual similarity between background characters and target speakers in the reference attention space, background regions may be erroneously assigned to the target speakers' attention regions. This results in them being driven by the corresponding speech signals, exhibiting undesired lip or facial motions.

To address this issue, we introduce additional bounding box annotations and model non-target character regions as independent categories during attention map estimation, which reduces the probability that they are absorbed into the target speaker regions. Nevertheless, attention level separation alone is insufficient to fully eliminate speech-driven motion. Since the original two-speaker MultiTalk formulation provides only two audio conditions, it does not assign an explicit audio condition to background regions. Therefore, when additional person boxes are available, we introduce an extra silent audio track as a dedicated background audio condition, mapping all non-target character regions to this silent condition. Consequently, as illustrated in Fig.~\ref{fig:multiti}, the audio cross-attention module precisely binds the two target speakers to their respective speech signals, while associating non-target characters with a silent condition. This mechanism effectively prevents the target speech from inducing unintended lip movements in background characters.

%% file: sec/4_training.tex
\section{Training}

The training pipeline of LongCat-Video-Avatar 1.5 consists of three 
progressive stages: Base Model Training, RLHF Training, 
and Acceleration Training.  
The first stage establishes the foundational capability of audio-driven 
avatar generation. The model is trained to synthesize temporally coherent 
and identity-preserving video conditioned on speech signals. 
The second stage leverages Reinforcement Learning from Human Feedback 
(RLHF)~\cite{ouyang2022training}, specifically employing Group Relative 
Policy Optimization (GRPO)~\cite{shao2024deepseekmath} to align the 
model's outputs with human preferences. This stage improves perceptual 
quality, expressiveness, and overall generation fidelity beyond what 
supervised training alone can achieve, ensuring that synthesized avatars 
are more natural and visually appealing to human observers.
The third stage focuses on optimizing inference efficiency. Through 
dedicated acceleration training, the model achieves high-quality avatar 
video synthesis at significantly reduced computational cost, enabling 
practical deployment without sacrificing generation quality.

\noindent\textbf{Base Model Training.}
We adopt the flow matching framework~\cite{lipman2022flow} for the 
generative process. Given a clean video latent $x_0$, a noise 
sample $\epsilon \sim \mathcal{N}(\mathbf{0}, \mathbf{I})$, 
and a timestep $t \in [0, 1]$, the noisy latent $x_t$ is 
constructed via linear interpolation:
\begin{equation}
    x_t = (1 - t) \cdot x_0 + t \cdot \epsilon.
    \label{eq:xt}
\end{equation}
The network is trained to predict the velocity 
$v_{\text{pred}}(x_t, c, t; \theta)$, 
where $c$ denotes the task conditions (\emph{e.g.}, text prompts, audios  
and conditional image/video latents), and $\theta$ denotes
the model parameters. The training objective minimizes MSE against
 ground truth velocity $v_t = x_0 - \epsilon$:
\begin{equation}
    \mathcal{L} = \mathbb{E}_{\epsilon, x_0, c, t}
    \left\| v_{\text{pred}}(x_t, c, t; 
    \theta) - v_t \right\|^2.
    \label{eq:loss}
\end{equation}
The base model training consists of multiple progressive stages,  as
outlined in Table~\ref{tab:training_stages}. the training begins with low-resolution pretraining, where the model 
learns the fundamental correspondence between speech signals and facial 
dynamics at a coarse spatial scale, establishing the core audio-driven 
generation capability.
Once the model demonstrates stable audio-driven capability, the training 
transitions to a high-resolution stage, enabling the model to synthesize 
fine-grained visual details and produce high-fidelity avatar videos with 
improved spatial quality.
A reference image module is subsequently introduced into the training 
pipeline, allowing the model to incorporate identity and appearance 
information from a given reference image. This stage equips the model 
with the ability to generate identity-preserving avatars conditioned on 
arbitrary reference inputs.
Finally, the model is trained on a large-scale multi-person dialogue 
dataset, extending its capability to handle multi-person conversational 
scenarios where multiple identities interact in a coherent and temporally 
consistent manner.

\noindent\textbf{RLHF Training.}
Following the base model training, we further improve the model 
performance through a post-training stage. Specifically, we adopt 
the GRPO method described in Section~\ref{sec:grpo}, incorporating 
multiple video quality-related reward signals to guide the optimization process.
We adopt most training hyperparameters and optimization settings from LongCat-Video~\cite{team2025longcat}. For the proposed multi-clip extension, we set the maximum rollout length to 5 clips and randomly sample the actual number of sequential clips during training. To keep training tractable, only the final clip contributes to GRPO optimization, while earlier clips are used solely as temporal context for subsequent rollout. For image-to-video, and video-continuation tasks, we further perform a first-frame hand-presence check with MediaPipe hand detection, which increases the proportion of hand-relevant training samples.

\noindent\textbf{Acceleration Training.}
We adopt the method described in Section~\ref{sec:fsg} for model distillation, 
enabling the distilled model to achieve generation quality comparable 
to 50-step inference using only 8 steps. During this stage, we observe 
that prolonged training leads to a noticeable degradation in visual 
realism. Therefore, we train the model for 400 steps, at which 
point the model achieves the best generation fidelity before 
quality decline sets in. Specifically, the generator learning rate is 
set to $2 \times 10^{-5}$, the fake score learning rate is set to 
$4 \times 10^{-6}$, and the update ratio between the generator and 
the fake scorer is set to $1\!:\!5$.

%% file: sec/5_evaluation.tex
\section{Evaluation}

\subsection{Settings}

We establish our human evaluation benchmark based on EvalTalker~\cite{zhou2025evaltalker}, which provides over 400 samples of varying difficulty. To further assess generalization, we supplement this with over 50 stylized images (e.g., cartoons and animals), yielding a total of 508 image-audio pairs. The benchmark encompasses diverse application scenarios (e.g., News Broadcasting, Education, Entertainment, Commercial), languages (Chinese/English), and visual styles (Realistic/Animated). It systematically categorizes difficulty across both audio dimensions (e.g., speaking speed, fluency, emotion, paralinguistics) and visual dimensions (e.g., person count, pose, background complexity, occlusion).

Following the quality assessment framework proposed by Zhou et al.~\cite{zhou2025evaltalker}, we adopt four perceptual quality dimensions for structured expert evaluation:

\begin{itemize}
    \item \textbf{Rationality}: Conformity with physical laws. This dimension assesses whether the generated subjects exhibit plausible body structures and natural movements, and whether background elements remain physically consistent without artifacts such as distorted limbs, unnatural object interactions, or garbled text.
    
    \item \textbf{Harmony}: Synergy among audio-visual elements. This dimension evaluates lip-audio synchronization, the harmony between facial expressions/body motions and speech content, and overall audio-visual coherence across subjects. Additionally, it assesses the visual naturalness of facial expressions and body movements from a purely visual perspective.
    
    \item \textbf{Stability}: Temporal consistency of image quality. This dimension captures degradations such as frame stuttering (jumpcuts), resolution or color tone fluctuations, blurring, and visual artifacts that disrupt smooth playback.
    
    \item \textbf{Consistency}: Identity preservation across the generated video. This dimension verifies that each subject maintains stable facial features, appearance attributes, and speaker identity throughout the sequence without mismatches or drifts.
\end{itemize}

Using this comprehensive benchmark, we evaluate our proposed model against seven state-of-the-art methods: LC-Video-Avatar 1.0~\cite{meituanlongcatteam2025longcatvideoavatartechnicalreport}, InfiniteTalk~\cite{yang2025infinitetalk}, OmniHuman-1.5~\cite{jiang2025omnihuman}, HeyGen~\cite{HeyGen}, Hedra~\cite{Hedra}, Kling Avatar 2.0~\cite{ding2025kling-avatar}, and OmniAvatar~\cite{gan2025omniavatar}. The standard task requires synthesizing a temporally coherent video from a single portrait and a driving audio clip. Note that in all the following evaluations LC-Video-Avatar 1.5 represents the accelerated model with 8 NFE (number of feedforward evaluation) for inference. 

We employ a dual-track evaluation methodology combining large-scale crowdsourced perception ratings with expert-level structured quality analysis:

\begin{itemize}
    \item \textbf{Subjective Track:} 770 crowdsourced evaluators rated each generated video on a 1--5 anthropomorphism scale, ultimately yielding 13,240 judgments.
    \item \textbf{Objective Track:} 10 domain experts conducted a structured quality analysis across four dimensions: Physical Rationality, Audio-Visual Harmony, Temporal Stability, and Identity Consistency, utilizing hierarchical problem taxonomies. To ensure enhanced precision, lip-synchronization was evaluated at $0.5\times$ playback speed.
    \item \textbf{Pairwise A/B Test:} We conducted a direct preference comparison between LC-Video-Avatar 1.5 and three leading commercial competitors to assess overall anthropomorphism.
\end{itemize}

\begin{figure}[t]
  \centering
  
  \begin{minipage}[b]{0.95\textwidth}
    \centering
    \includegraphics[width=\linewidth]{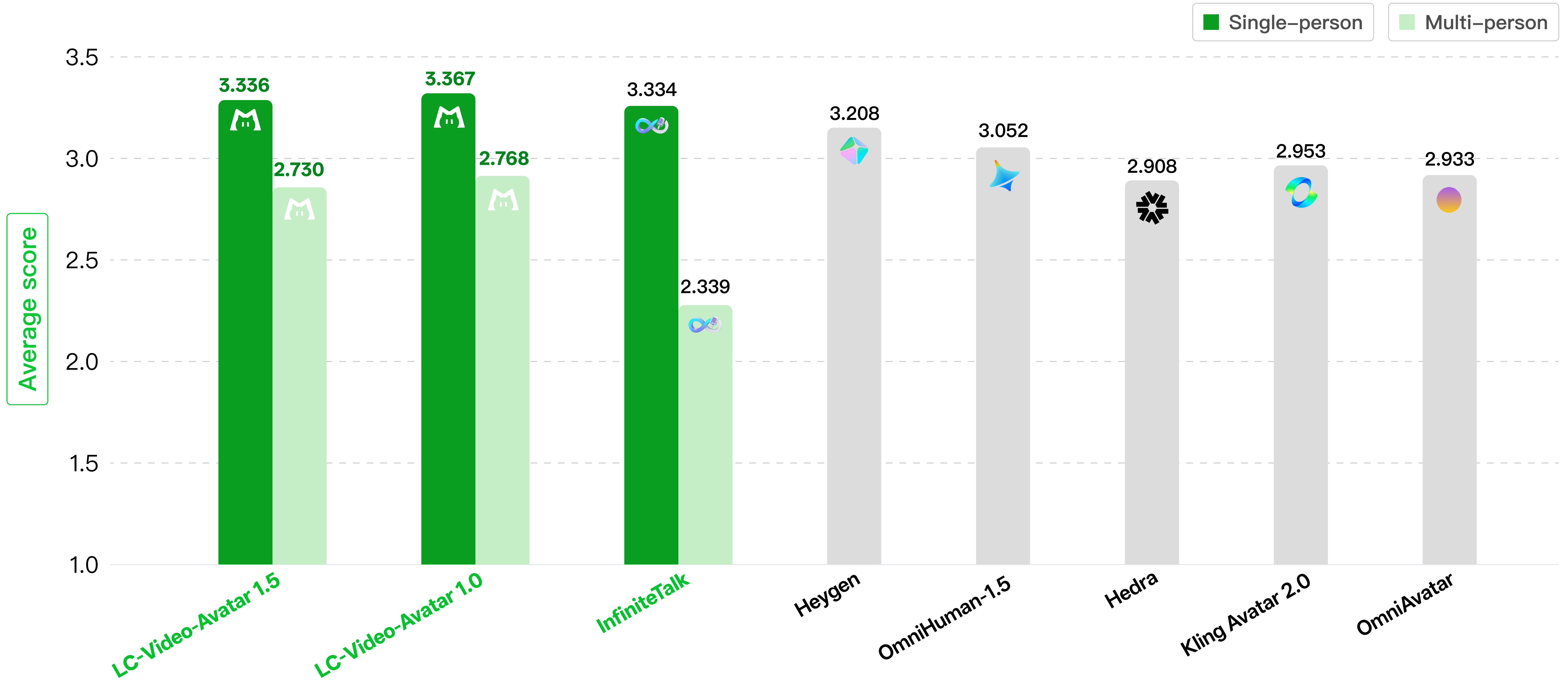}
    \caption{Human-likeness comparison across different methods in single person talking and multiple person conversation.}
    \label{fig:single_and_multi}
  \end{minipage}
\end{figure}

\begin{figure}[htbp]
  \centering
  
  \begin{minipage}[t]{0.48\textwidth}
    \centering
    \includegraphics[width=\linewidth]{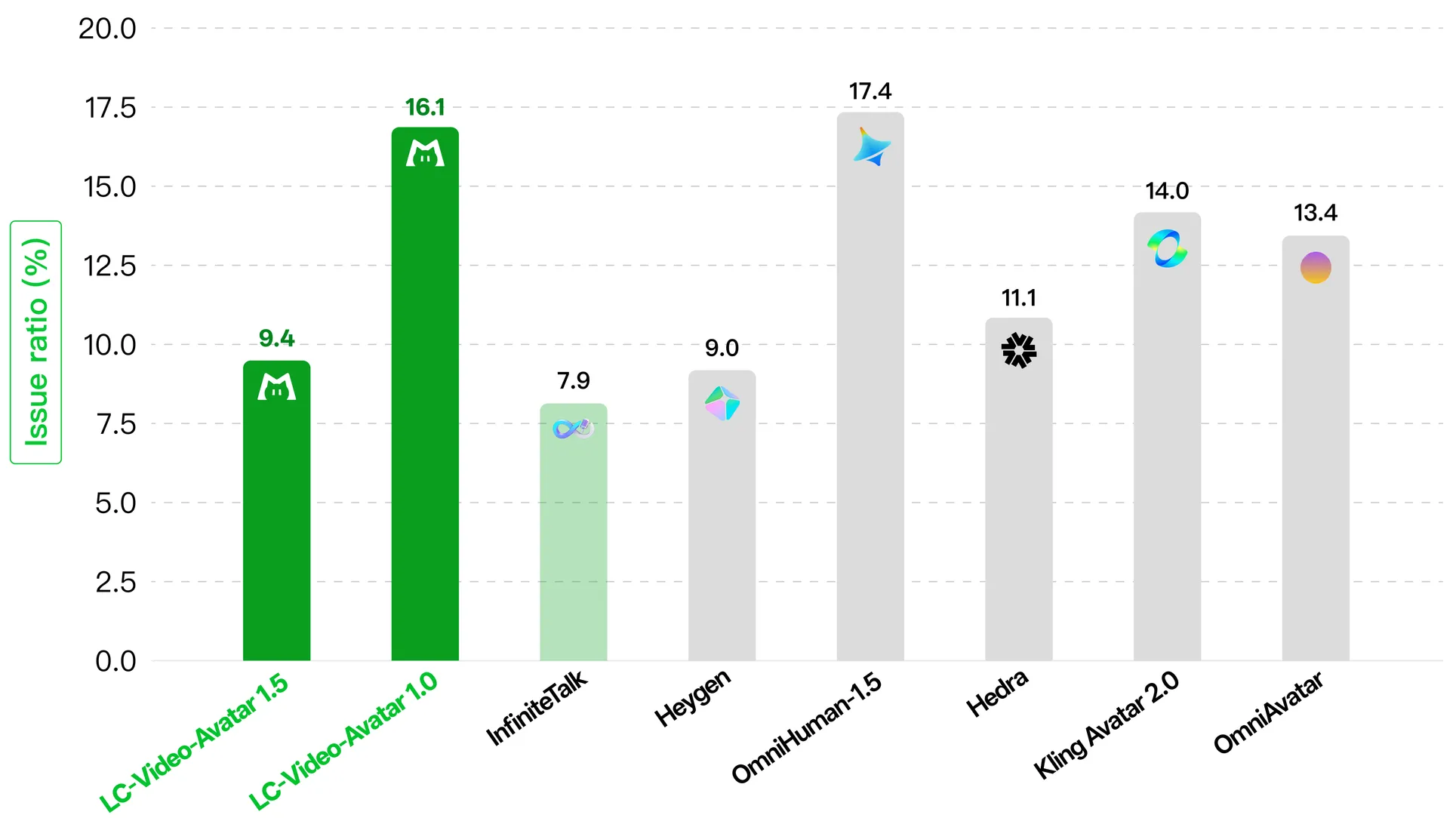}
    \caption{Background distortion in rationality.}
    \label{fig:background_distortion}
  \end{minipage}
  \hfill
  \begin{minipage}[t]{0.48\textwidth}
    \centering
    \includegraphics[width=\linewidth]{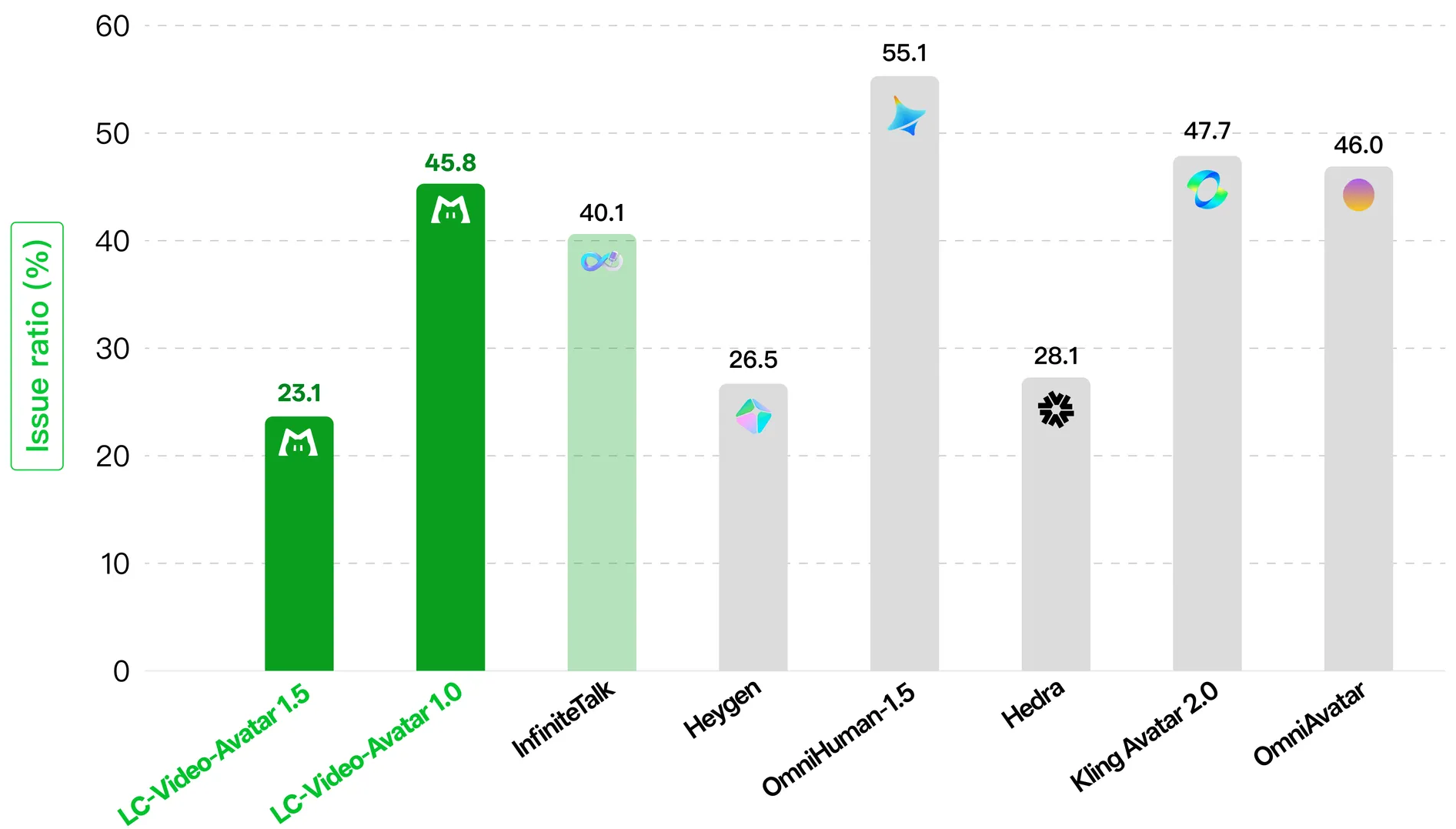}
    \caption{Subject distortion in rationality.}
    \label{fig:subject_distortion}
  \end{minipage}
  
  \vspace{0.5cm} 
  
  \begin{minipage}[t]{0.48\textwidth}
    \centering
    \includegraphics[width=\linewidth]{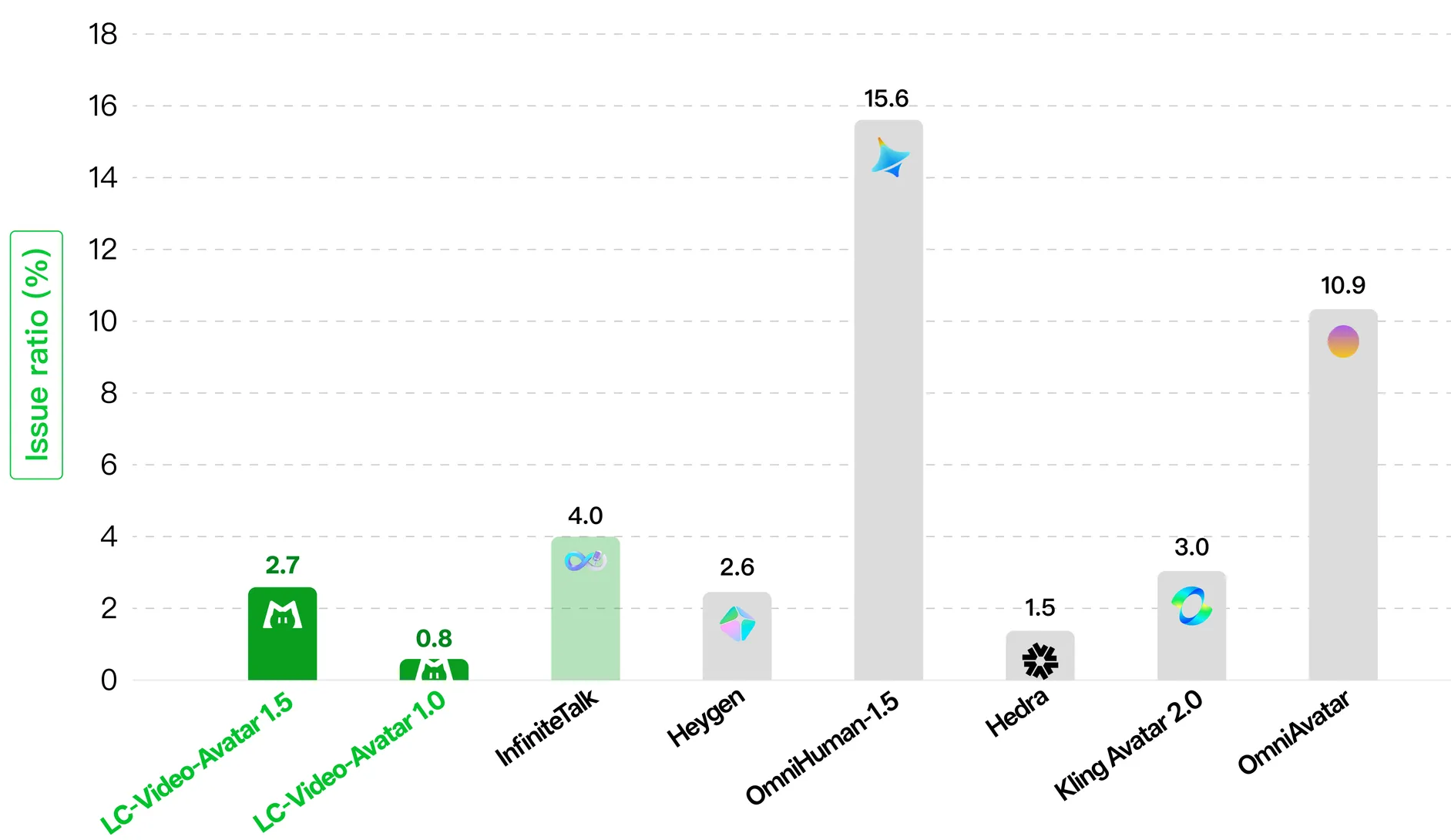}
    \caption{Tone error accumulation in stability.}
    \label{fig:tone_error_accumulation}
  \end{minipage}
  \hfill
  \begin{minipage}[t]{0.48\textwidth}
    \centering
    \includegraphics[width=\linewidth]{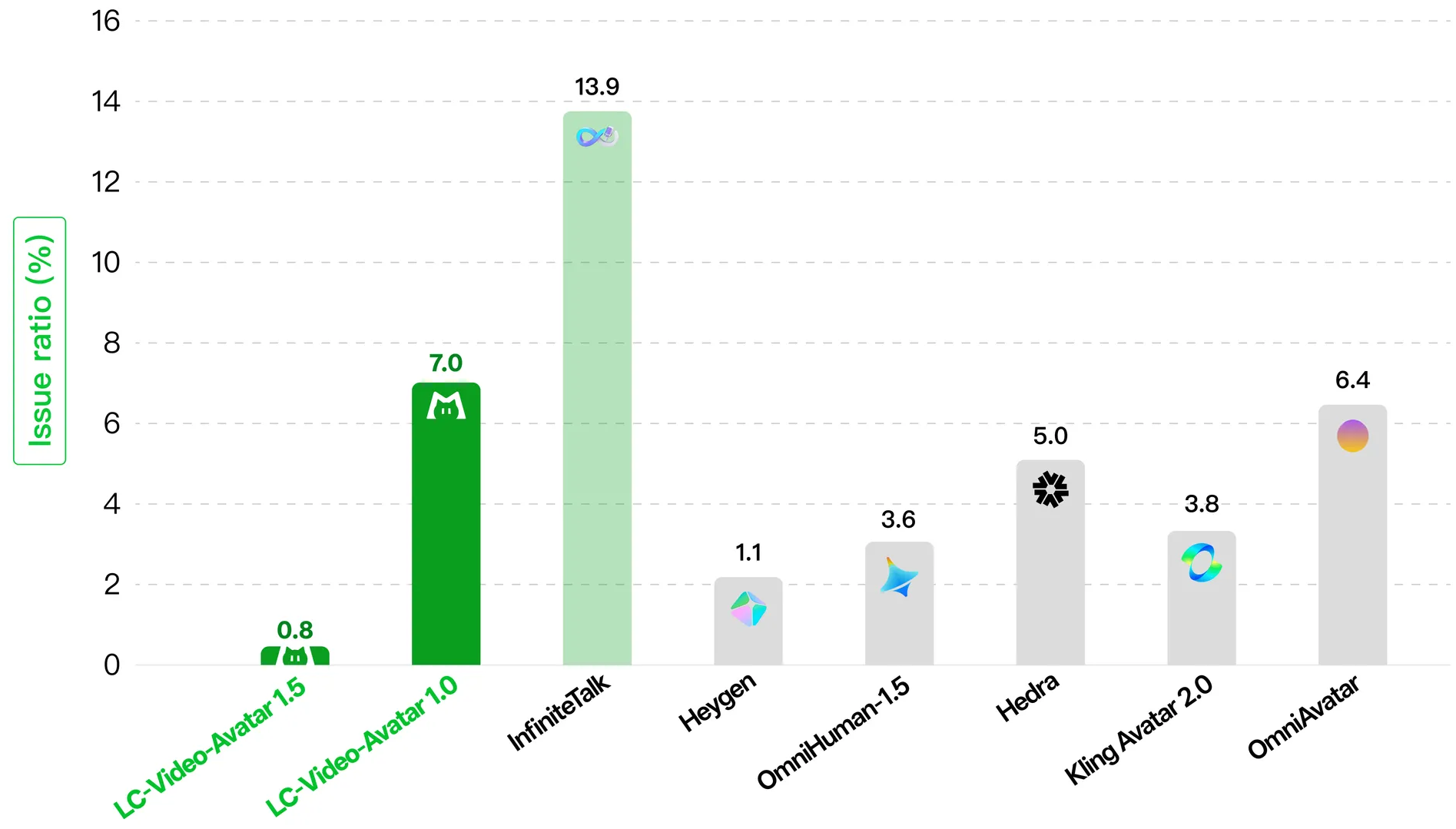}
    \caption{Frame jumpcut in stability.}
    \label{fig:jumpcut}
  \end{minipage}

  \vspace{0.5cm}
  
  \begin{minipage}[t]{0.48\textwidth}
    \centering
    \includegraphics[width=\linewidth]{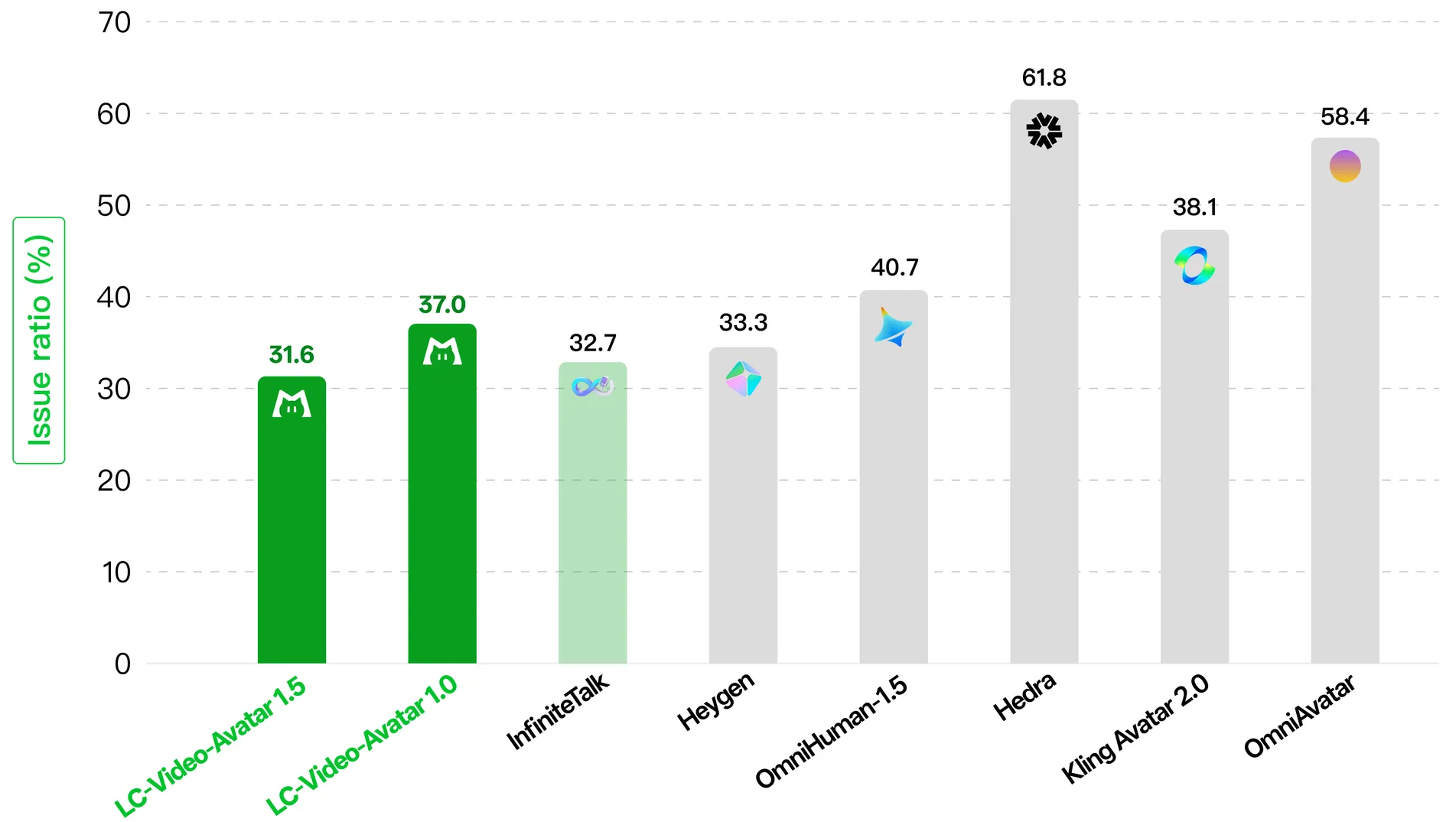}
    \caption{Lip synchronization in harmony.}
    \label{fig:lip_synchronization}
  \end{minipage}
  \hfill
  \begin{minipage}[t]{0.48\textwidth}
    \centering
    \includegraphics[width=\linewidth]{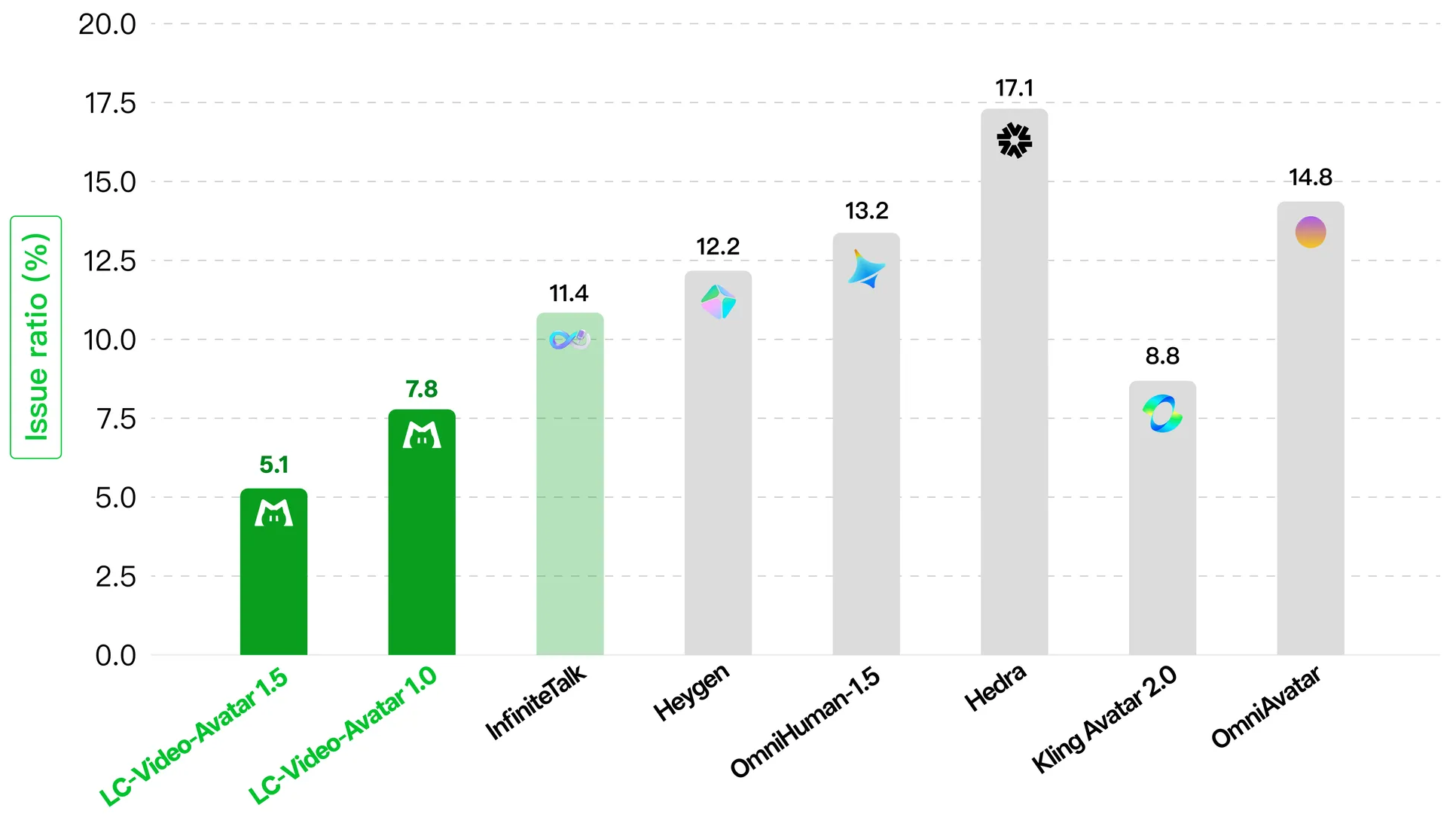}
    \caption{Face/body synchronization in harmony.}
    \label{fig:face_body_synchronization}
  \end{minipage}

\vspace{0.5cm} 

  \begin{minipage}[t]{0.48\textwidth}
    \centering
    \includegraphics[width=\linewidth]{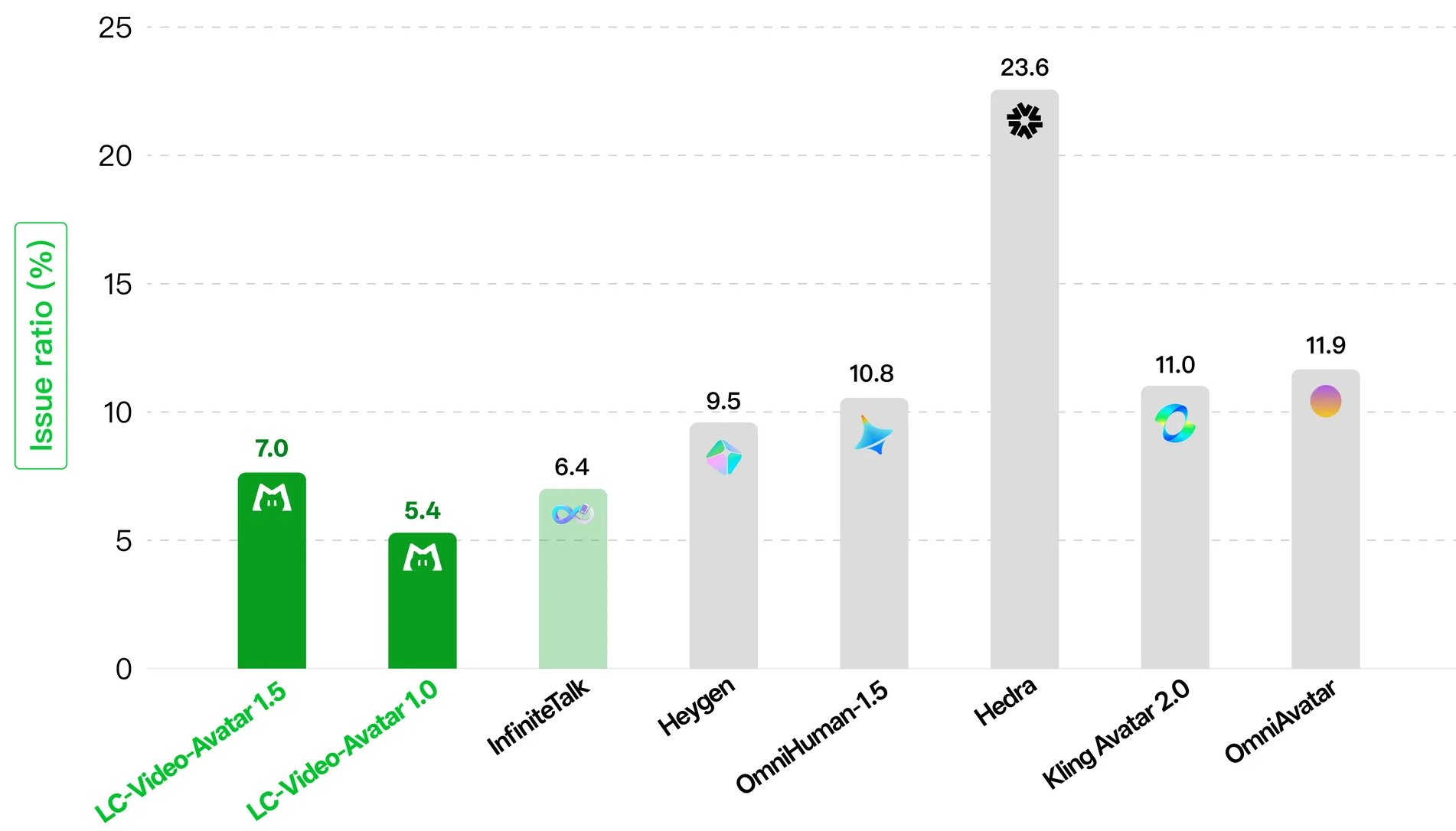}
    \caption{Body naturalness in harmony.}
    \label{fig:body_naturalness}
  \end{minipage}
  \hfill
  \begin{minipage}[t]{0.48\textwidth}
    \centering
    \includegraphics[width=\linewidth]{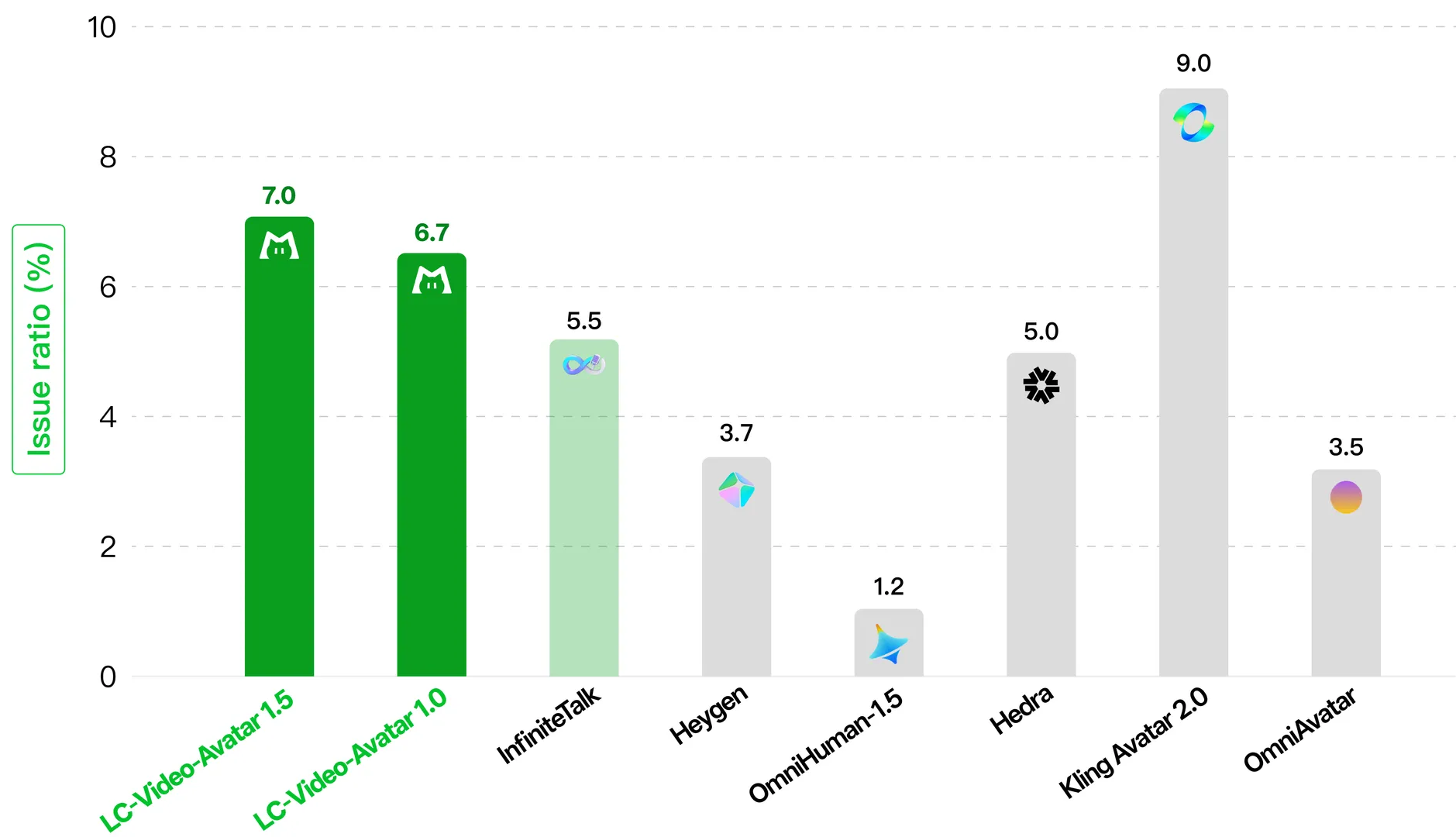}
    \caption{Facial expression naturalness in harmony.}
    \label{fig:emtoin_naturalness}
  \end{minipage}

\end{figure}

\begin{figure*}[t]
  \centering
  \begin{minipage}[t]{0.95\textwidth}
    \centering
    \includegraphics[width=\linewidth]{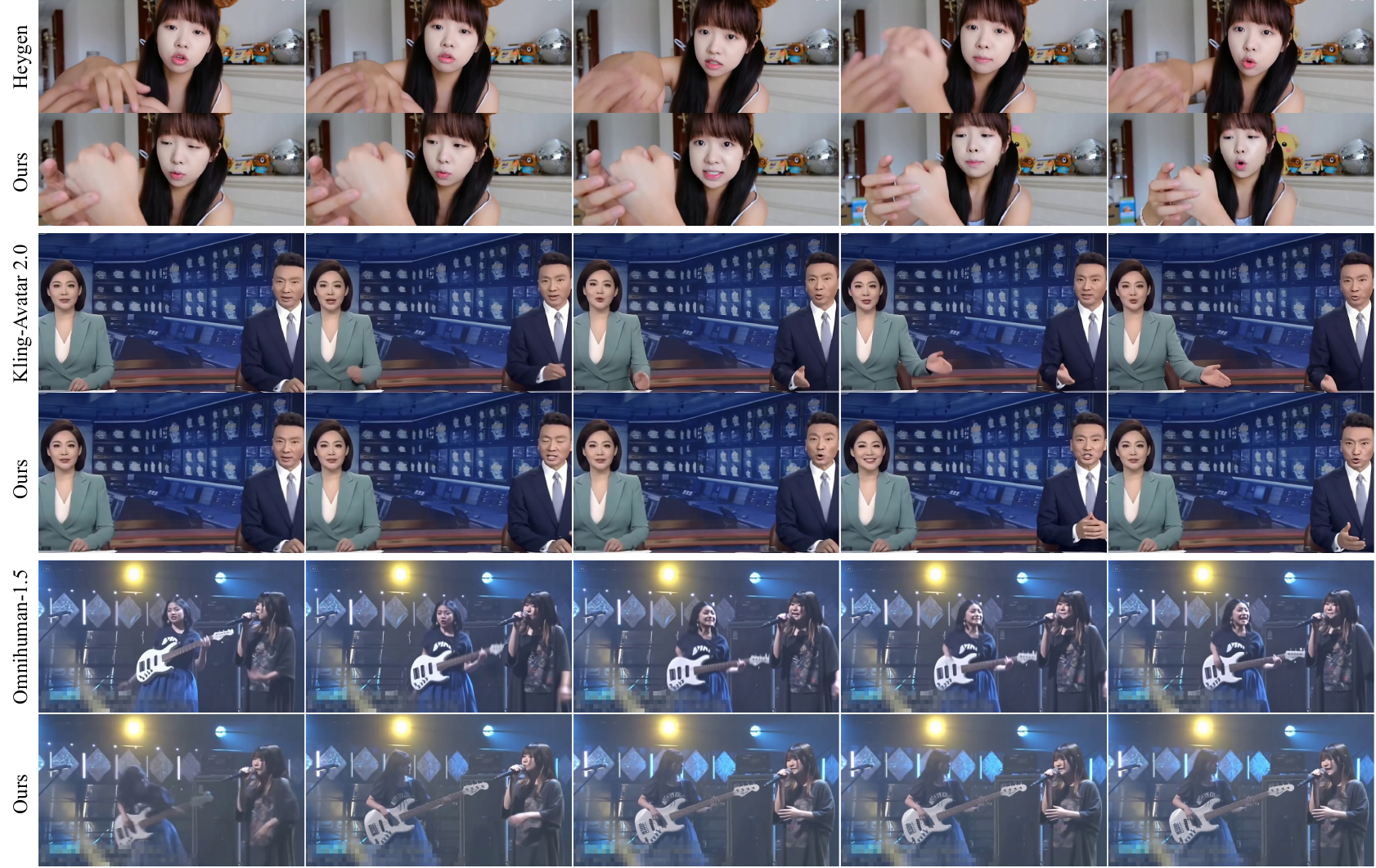}
    \caption{Visual comparison in rationality.}
    \label{fig:vis_rationality}
  \end{minipage}
  \vspace{0.5cm}
  \begin{minipage}[t]{0.95\textwidth}
    \centering
    \includegraphics[width=\linewidth]{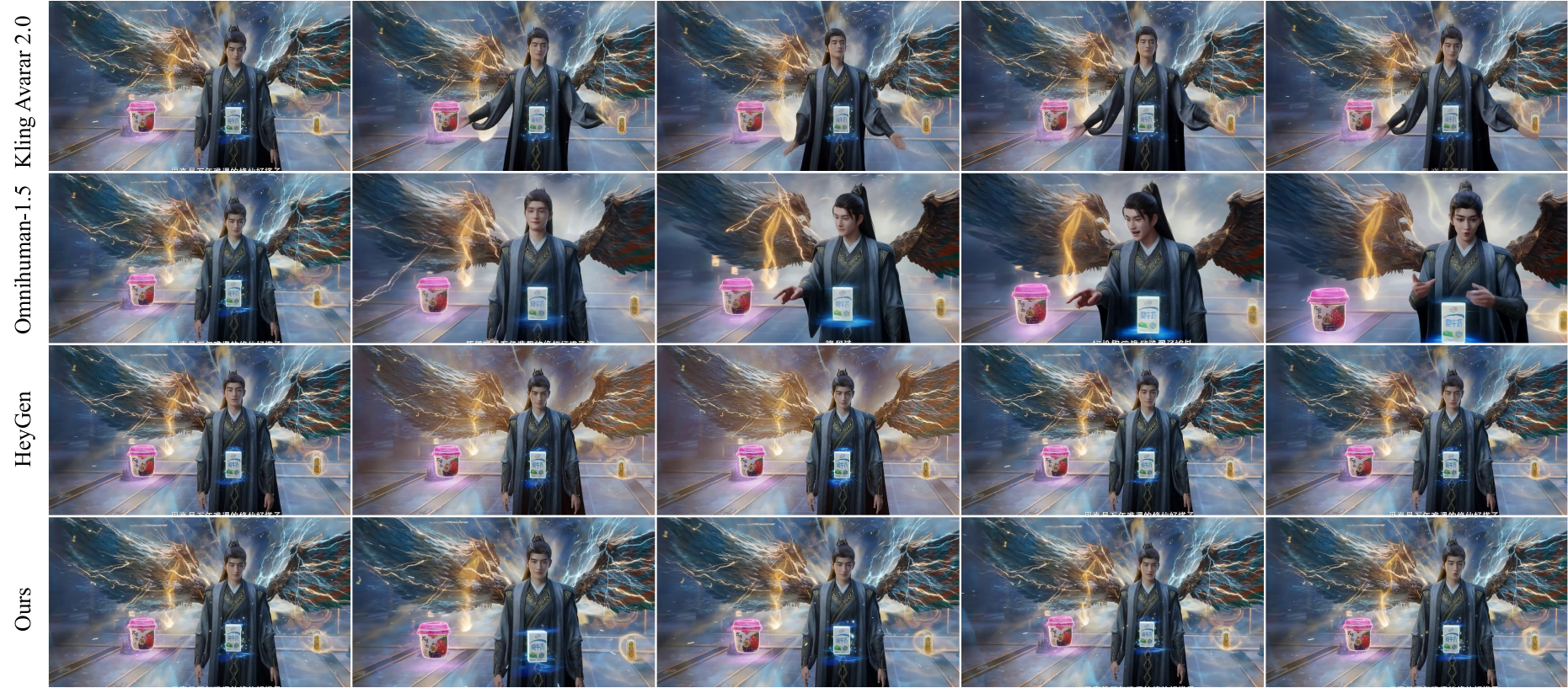}
    \caption{Visual comparison in stability.}
    \label{fig:stable}
  \end{minipage}
\end{figure*}

\begin{figure*}[t]
  \centering
  \begin{minipage}[t]{0.95\textwidth}
    \centering
    \includegraphics[width=\linewidth]{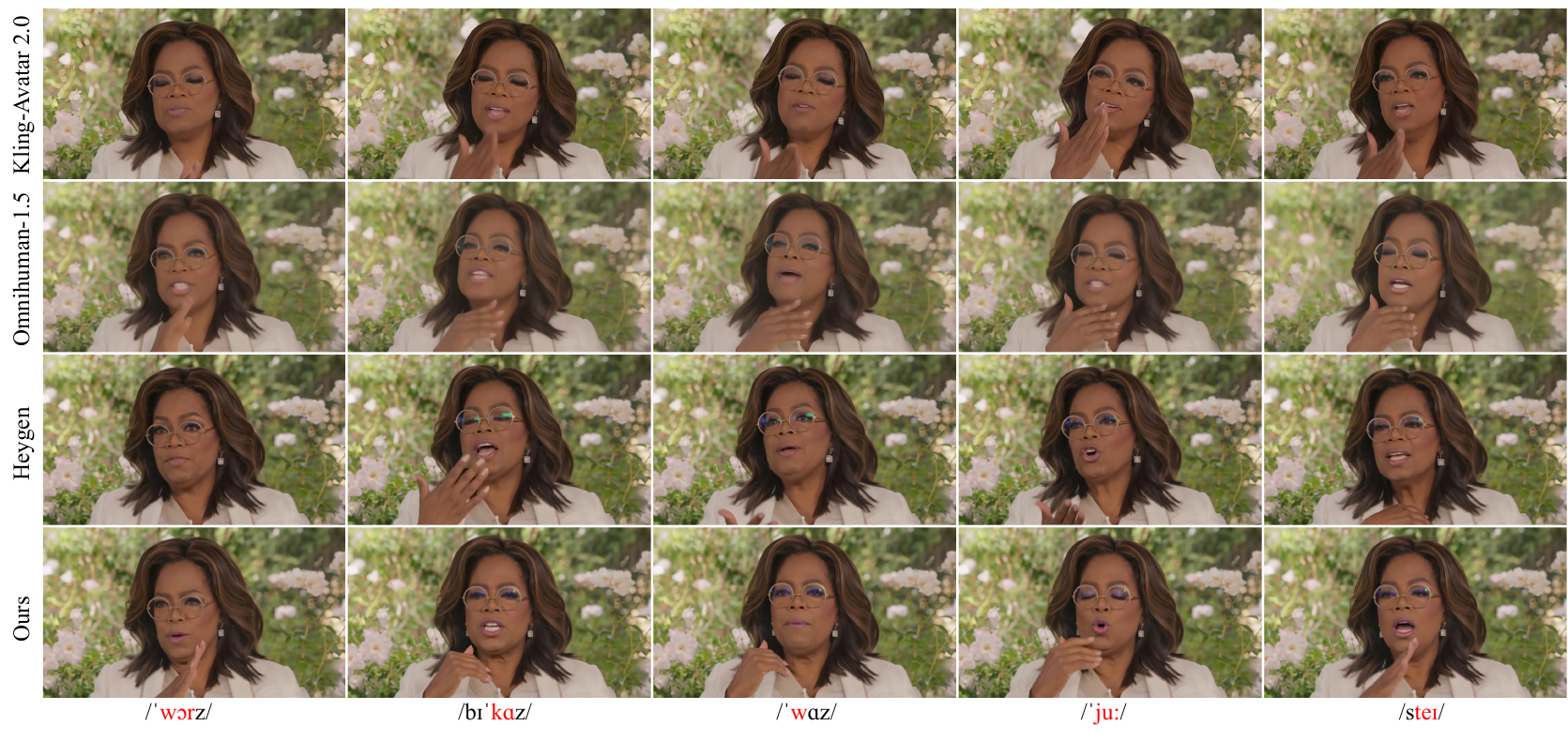}
    \caption{Visual comparison in talking head scenarios.}
    \label{fig:koubo}
  \end{minipage}
  \vspace{0.5cm} 
  \begin{minipage}[t]{0.95\textwidth}
    \centering
    \includegraphics[width=\linewidth]{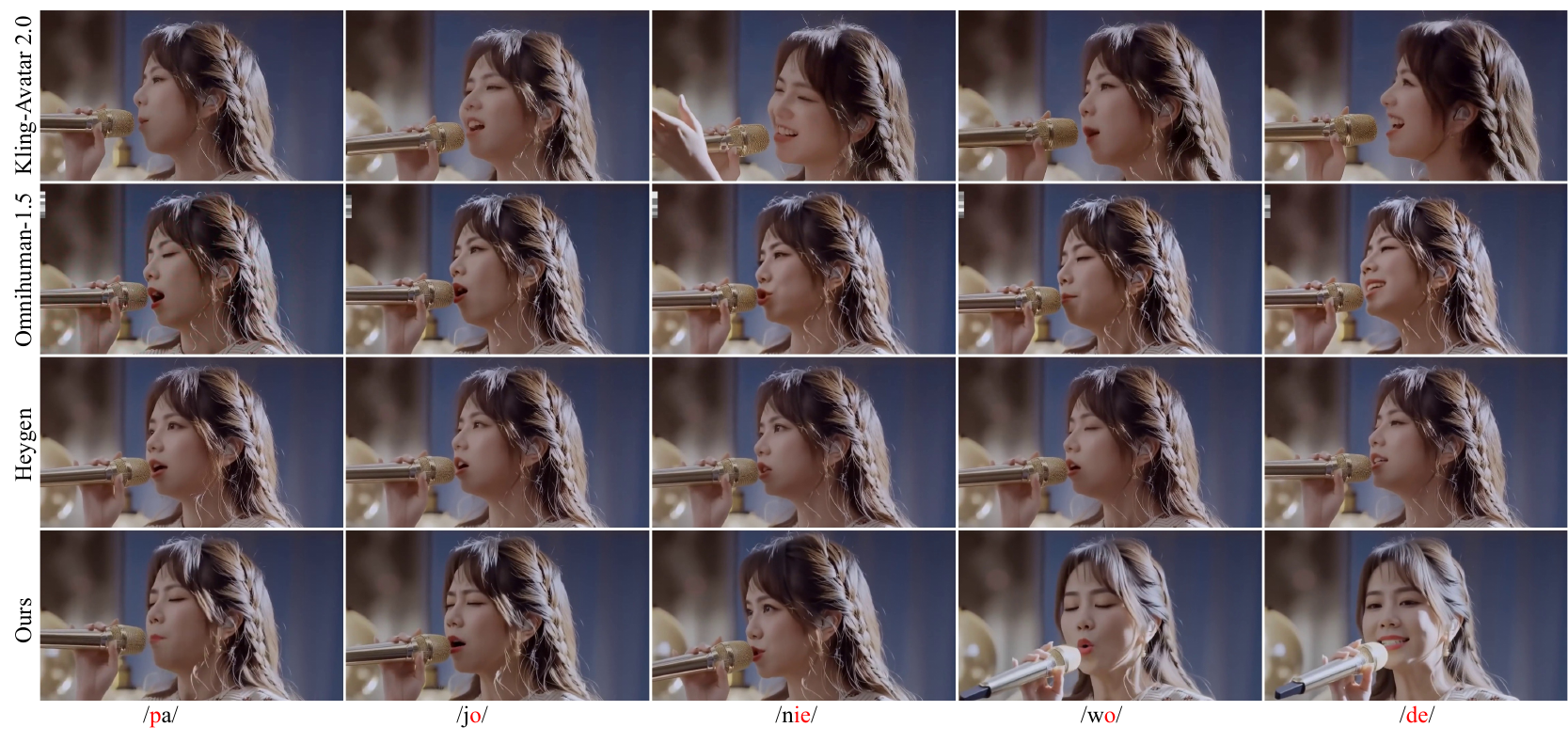}
    \caption{Visual comparison in music scenarios.}
    \label{fig:music}
  \end{minipage}
\end{figure*}

\begin{figure*}[t]
  \centering
  \begin{minipage}[t]{0.95\textwidth}
    \centering
    \includegraphics[width=\linewidth]{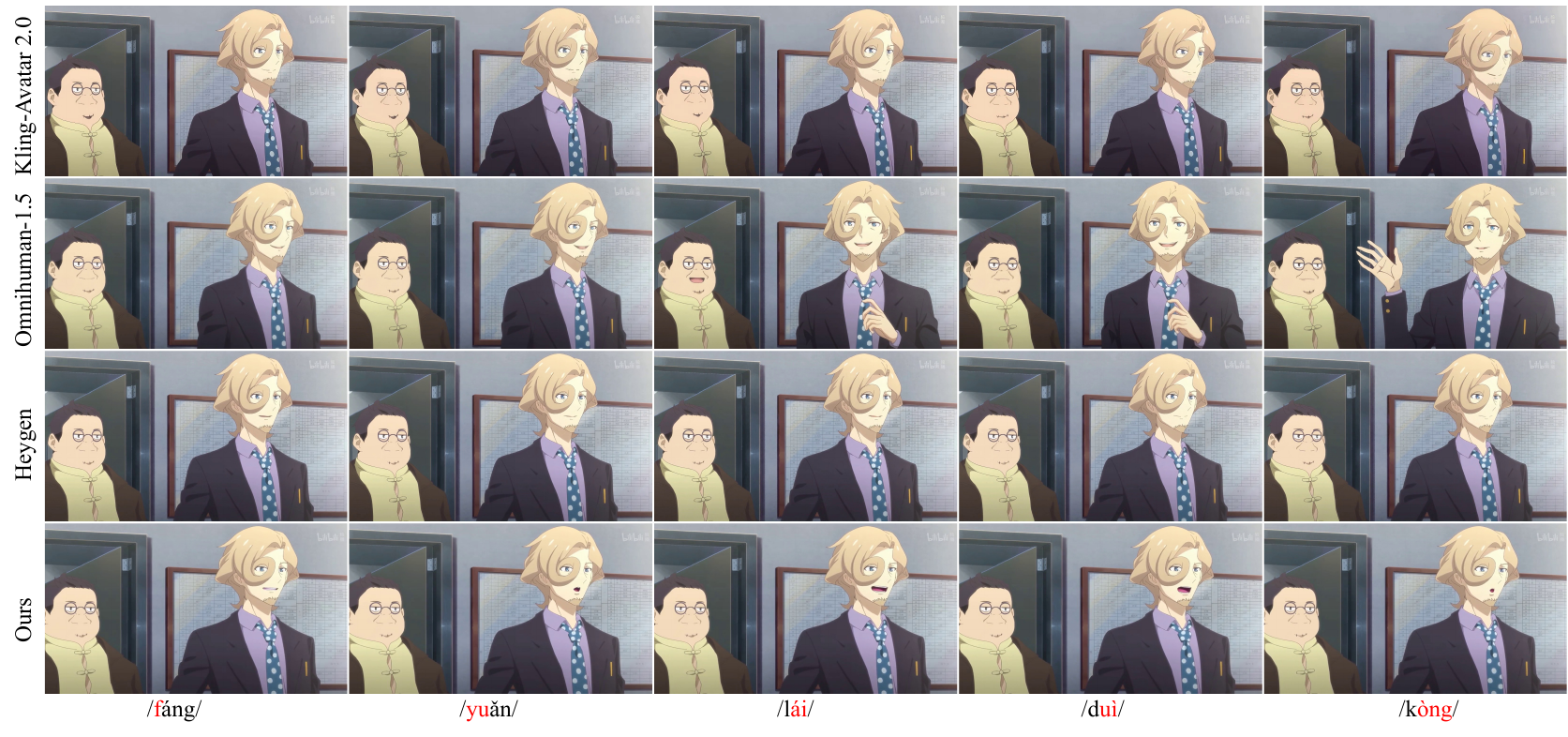}
    \caption{Visual comparison in anime scenarios.}
    \label{fig:dongman}
  \end{minipage}
  \vspace{0.5cm} 
  \begin{minipage}[t]{0.95\textwidth}
    \centering
    \includegraphics[width=\linewidth]{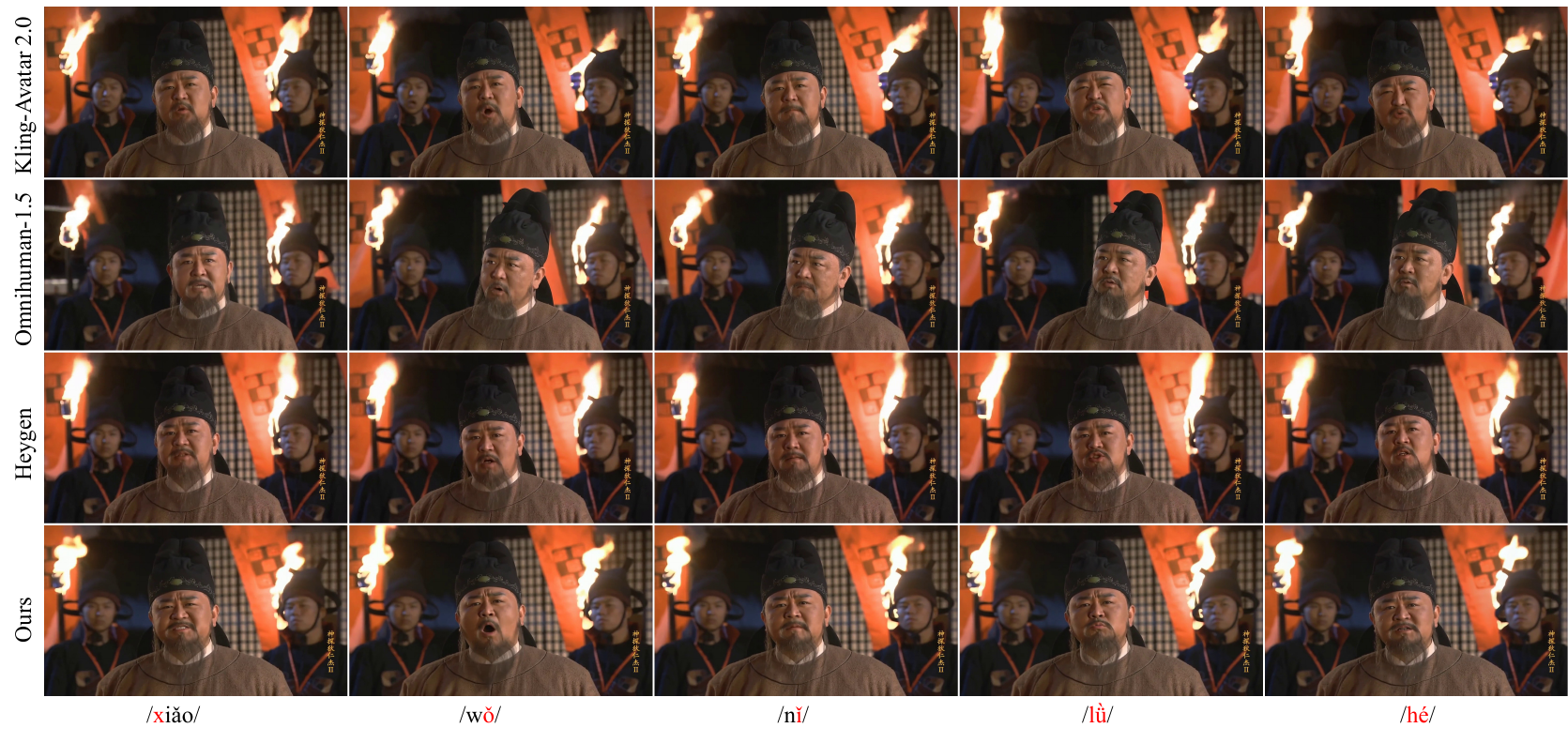}
    \caption{Visual comparison in performance scenarios.}
    \label{fig:biaoyan}
  \end{minipage}
\end{figure*}

\begin{figure*}[t]
  \centering
  \includegraphics[width=0.95\linewidth]{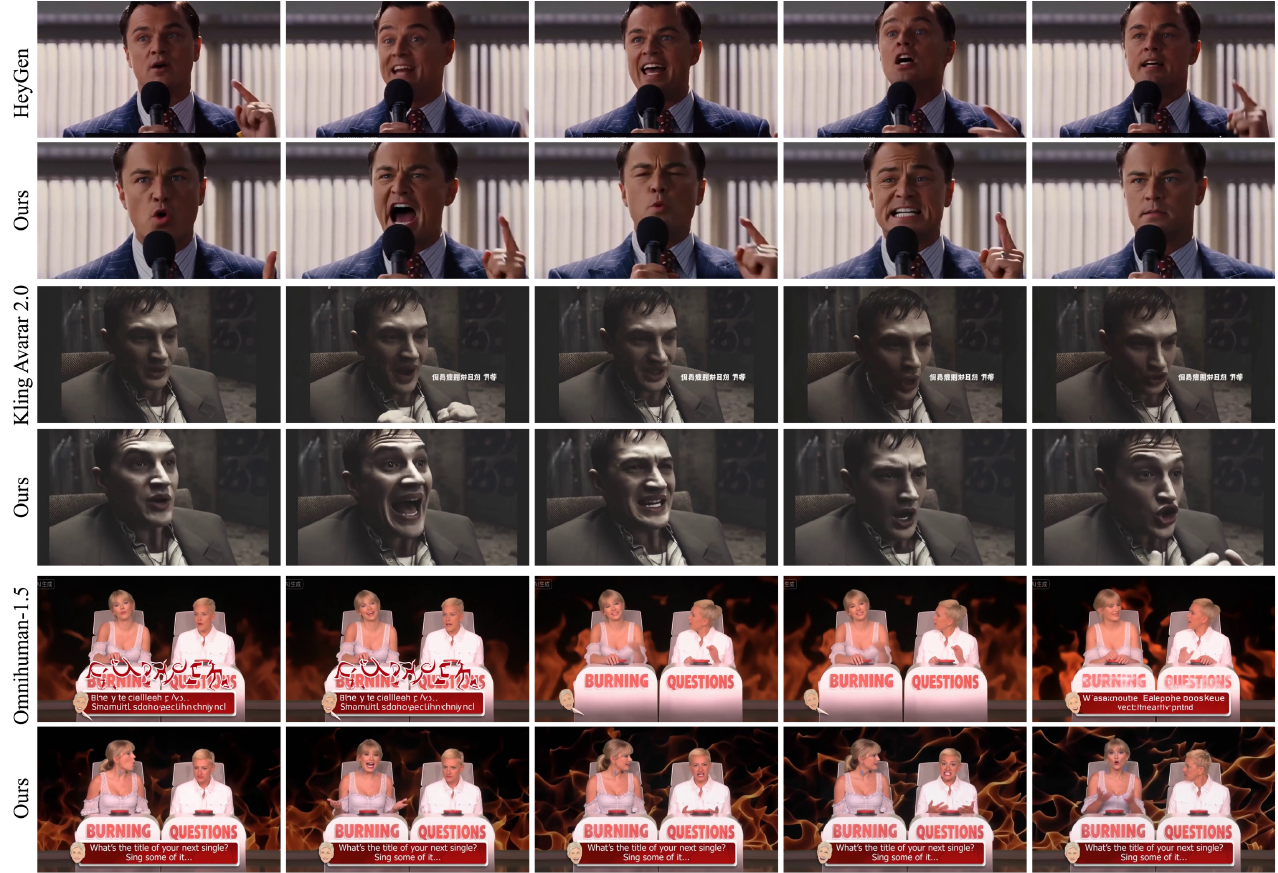}
  \caption{Visual comparison in emotional expression scenarios.}
  \label{fig:qinggan}
\end{figure*}

\subsection{Overall Human-likeness Evaluation}

To thoroughly investigate the capabilities of current virtual human generation methods, we designed a comprehensive evaluation covering both single-person and multi-person scenarios. Our evaluation protocol is tailored to the supported features of each method: models capable of multi-person generation were assessed in both configurations, whereas those lacking multi-person support were evaluated exclusively on the single-person subset.

As illustrated in Fig.~\ref{fig:single_and_multi}, the comparative analysis reveals distinct tiers in human-likeness. In the single-person setting, the top three methods—LC-video-avatar 1.5, LC-video-avatar 1.0, and InfiniteTalk—demonstrate comparable leading performance, which is closely followed by Heygen and OmniHuman-1.5. However, the multi-person task introduces greater complexity and shifts the performance dynamics. Among the methods supporting multi-person synthesis, the two LC-video-avatar variants maintain similar levels of human-likeness, both significantly outperforming the third model, InfiniteTalk.

Despite the progress made by these state-of-the-art methods, overall trends indicate a broader challenge: current virtual human models still face a considerable gap before achieving highly realistic human-likeness. Further analysis indicates that this perceptual gap is primarily attributable to two factors: the first is a deficiency in physical rationality, where models often generate physically implausible movements, anatomical distortions, or unnatural interactions (see details in Sec.~\ref{sec:exp-eval}). The second major factor is suboptimal audio-visual synchronization, which breaks the illusion of natural speech and heavily degrades the overall perceptual quality.

\subsection{Expert-level Objective Quality Evaluation} \label{sec:exp-eval}

Furthermore, we conduct an expert-level objective quality analysis across four complementary perceptual dimensions: temporal stability, physical rationality, identity consistency, and harmony (\textit{i.e.,} audio-visual harmony). This decomposed evaluation enables the precise identification of strengths and remaining challenges. As illustrated in the radar chart in Fig.~\ref{fig:comp1}, where performance is quantified as ($100 - \text{Issue Rate}$) (higher is better), LC-Video-Avatar 1.5 achieves industry-leading stability and rationality, alongside state-of-the-art identity consistency. However, audio-visual harmony remains an open challenge across the entire field. To provide a more granular breakdown of these results, we calculate the specific issue rates across the evaluated models.

\paragraph{Rationality.}
Rationality evaluates whether the synthesized avatar's movements, expressions, and environmental interactions comply with real-world physical laws and biomechanics, encompassing aspects like subject and background distortion. Figs.~\ref{fig:background_distortion} and \ref{fig:subject_distortion} detail the issue rates of these artifacts across various methods, revealing that physical rationality remains a prevalent bottleneck for current generative models, especially subject distortion. LC-Video-Avatar 1.5 achieves a leading performance in this area. This enhanced structural rationality is primarily attributed to the integration of GRPO during training. By employing reward signals that explicitly penalize unnatural or physically incorrect generations, GRPO effectively guides the network to produce highly rational and physically grounded virtual humans. 
Besides, we observe that DMD distillation also contributes to rationality improvements, particularly in reducing hand distortion and suppressing exaggerated facial expressions.
Fig. \ref{fig:vis_rationality} provides a comprehensive qualitative visual comparison with other methods. As observed, both Kling-Avatar 2.0 and Heygen struggle with fine-grained hand generation, exhibiting severe structural deformations in the hand regions. Meanwhile, Omnihuman-1.5 suffers from severe depth ordering and occlusion failures; specifically, an arm initially occluded behind the guitar abruptly and unnaturally shifts to the foreground, overlapping the instrument.

\paragraph{Stability.}
In terms of temporal stability, the evaluation primarily encompasses frame jump cuts, color tone error accumulation, and resolution shifts. Fig.~\ref{fig:tone_error_accumulation} and \ref{fig:jumpcut} illustrate the issue rates specifically for tone error accumulation and frame jump cuts, respectively. We analyze the performance across Tone Error Accumulation and Frame Jumpcut metrics. Regarding Tone Error Accumulation, OmniHuman 1.5 exhibits substantial error build-up. This limitation is potentially tied to its Pseudo Frame mechanism, which appears insufficient to completely prevent the progressive accumulation of visual artifacts over time. In contrast, our approach inherits the reference skip attention mechanism from our 1.0 architecture, proving its continued efficacy in suppressing error propagation. It is worth noting that while our 1.5 model demonstrates a marginally higher error rate than our 1.0 baseline, this is a deliberate and calculated trade-off. The integration of DMD2 distillation in version 1.5 substantially accelerates inference speed, which inevitably introduces a minor compromise in the raw generation quality. Furthermore, in terms of Frame Jumpcut, our proposed method achieves the lowest occurrence rate among all compared models. This strong temporal consistency can be attributed to our data processing pipeline, which incorporates a specialized operator explicitly designed for jumpcut detection and filtering. This result indicates that careful data curation and preprocessing contribute meaningfully to the temporal stability and seamlessness of synthesized avatar videos.
Beyond structural identity, temporal visual stability is another crucial aspect of consistency. As shown in Fig.\ref{fig:stable}, competing methods frequently suffer from noticeable color shifts across frames. In contrast, our proposed method maintains a highly stable color profile and consistent illumination throughout the entire video sequence. This absence of color flickering and degradation indicates the temporal stability and robustness of our method.

\paragraph{Harmony.}
Regarding audio-visual harmony, our evaluation covers key dimensions such as lip-sync, expression-motion alignment, and face-body synchronization. Figs.~\ref{fig:lip_synchronization}, ~\ref{fig:face_body_synchronization} and \ref{fig:body_naturalness} detail the issue rates for face-body and lip synchronization, respectively. 
In Fig.~\ref{fig:emtoin_naturalness}, we present an evaluation focused strictly on the visual naturalness of the generated avatars. Although the evaluated videos contain audio, human evaluators were instructed to base their assessments solely on visual cues, specifically measuring the issue ratios (where a lower score indicates better performance) of unnatural body movements and facial expressions. In this context, "unnaturalness" is characterized by implausible motion trajectories. This includes evaluating whether the avatar's body and face maintain natural, subtle dynamics even during silent pauses (when the speaker is not talking), as well as identifying visual artifacts such as localized freezing, micro-jittering, or excessive and erratic shaking of the face. Based purely on these visual criteria, the models exhibit distinct comparative strengths. For body naturalness, LC-video-avatar 1.0 achieves the best performance, demonstrating highly stable and reasonable body kinematics. It is closely followed by InfiniteTalk and LC-video-avatar 1.5, which also show strong capabilities in maintaining bodily harmony. Conversely, regarding the naturalness of facial expressions, OmniHuman-1.5 stands out as the top performer. It excels in minimizing facial artifacts, ensuring the most stable and fluid facial muscle dynamics among all evaluated methods.
Compared to the v1.0 baseline, v1.5 demonstrates a consistent reduction in issue rates across both metrics, reflecting a notable enhancement in motion naturalness and audio-visual harmony. We attribute this improvement to the architectural upgrade of our audio feature extraction module: replacing the Wav2vec encoder with the contextually robust Whisper-large model. This transition allows v1.5 to capture richer phonetic and prosodic representations, thereby facilitating tighter temporal alignment between driving audio signals and facial/bodily dynamics. Additionally, qualitative results (Figs.~\ref{fig:koubo}, \ref{fig:music}, \ref{fig:dongman}, \ref{fig:biaoyan}, and \ref{fig:qinggan}) demonstrate that our approach consistently achieves superior lip synchronization across diverse scenarios, including talking head, music, anime, and performance. Our model also demonstrates improved expressiveness in emotion-driven scenarios, as illustrated in~\ref{fig:qinggan}.
 
\paragraph{Consistency.}
The consistency measures whether the identity changes in the generated video. As shown in Fig.\ref{fig:comp1}, LC-Video-Avatar 1.5 performs the best in identity preservation, followed by LC-Video-Avatar 1.0, InfiniteTalk, Hedra, Heygen, and Kling Avatar 2.0. In contrast, OmniHuman 1.5 and OmniAvatar show weaker identity preservation.

We also conduct pairwise A/B preference tests to measure holistic perceptual quality against leading commercial systems. In each trial, evaluators view two anonymized videos generated from identical inputs and select their preferred result based on overall human-likeness (Fig. \ref{fig:comp2}). This protocol directly captures end-user preference without decomposition bias.
LC-Video-Avatar 1.5 achieves majority preference against all three competitors, with the most decisive advantage over Kling Avatar 2.0 , followed by OmniHuman-1.5 and Heygen. These results indicate that LC-Video-Avatar 1.5 achieves competitive or superior preference against all evaluated commercial alternatives.

\begin{table}[t]
\centering
\caption{Comparison between Base and Fast variants. Higher human-likeness scores are better, while lower issue rates are better for the remaining metrics.}
\label{tab:base-fast-comparison}
\resizebox{.8\columnwidth}{!}{%
\begin{tabular}{lcccccc}
\toprule
\textbf{Method} 
& \textbf{Human-likeness} $\uparrow$
& \textbf{Human-likeness} $\uparrow$
& \textbf{Rationality}
& \textbf{Harmony}
& \textbf{Stability}
& \textbf{Consistency} \\
& \textbf{score (single)}
& \textbf{score (multi)}
& \textbf{issue rate} $\downarrow$
& \textbf{issue rate} $\downarrow$
& \textbf{issue rate} $\downarrow$
& \textbf{issue rate} $\downarrow$ \\
\midrule
Base 
& \textbf{3.389}
& 2.676
& 51.5
& \textbf{44.2}
& 12.3
& 6.2 \\
Fast 
& 3.336
& \textbf{2.730}
& \textbf{32.4}
& 45.0
& \textbf{4.3}
& \textbf{5.9} \\
\bottomrule
\end{tabular}%
}
\end{table}

\begin{figure*}[t]
  \centering
  \includegraphics[width=0.99\linewidth]{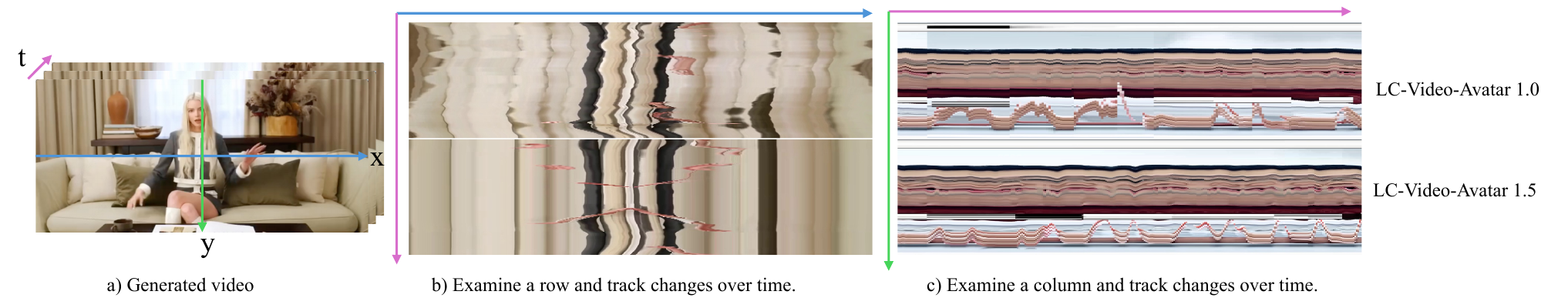}
  \caption{Stability comparison with LC-Video-Avatar 1.0.}
  \label{fig:temporal_profiles}
\end{figure*}

\begin{figure*}[t]
  \centering
  \includegraphics[width=0.99\linewidth]{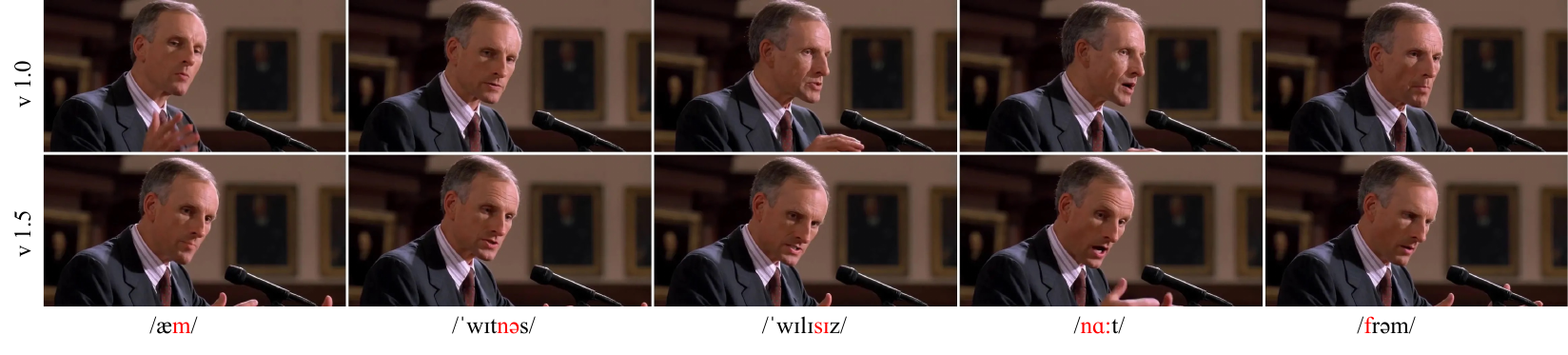}
  \caption{Lip synchronization comparison between v1.0 and v1.5.}
  \label{fig:lip-selfcomp}
\end{figure*}

\paragraph{Comparison with LC-Video-Avatar 1.0} We compare LC-Video-Avatar 1.5 with v1.0 in stability and lip synchronization.
To intuitively illustrate the temporal dynamics, we follow \cite{zhou2024upscale} and employ spatiotemporal slice visualization in Fig. \ref{fig:temporal_profiles}(b) and \ref{fig:temporal_profiles}(c). Specifically, these temporal profiles are generated by sampling a fixed spatial cross-section (either along the X-axis or Y-axis) across all video frames and concatenating them along the time (T) axis. This technique captures the temporal evolution of the selected slice, where smooth and continuous textures indicate high temporal consistency, whereas discontinuities reveal frame drops or jitter. As shown in the left panel, LC-Video-Avatar 1.5 achieves greater camera stability. Furthermore, the right panel highlights its enhanced temporal consistency, making it significantly less prone to frame skipping compared to version 1.0. Beyond temporal stability, we also evaluate lip synchronization capabilities. As illustrated in Fig. \ref{fig:lip-selfcomp}, version 1.5 demonstrates highly precise mouth dynamics and tighter audio-lip alignment than its predecessor.

\subsection{Comparison between the Basic and the Accelerated Version}
The model denoted as LC-Video-Avatar 1.5 in our evaluations refers to the accelerated version, which requires only 8 forward evaluations (i.e., 8 NFEs). Here, we compare its performance with the Base model, which performs a 50-step inference process requiring 3 forward passes per step, culminating in a total of 150 NFEs. As shown in Table. \ref{tab:base-fast-comparison}, the comparison reveals a distinct trade-off between expressive richness and generation stability. The Base model maintains a noticeable advantage in overall human-likeness and lip synchronization. Furthermore, it produces greater motion diversity, more nuanced facial expressions, and richer camera dynamics. Conversely, the accelerated LC-Video-Avatar 1.5 excels in maintaining visual stability, demonstrating significantly lower distortion rates across critical regions such as the hands, body, and face.

%% file: sec/6_conclusion.tex
\section{Conclusion and Future Work}

In this report, we presented LongCat-Video-Avatar 1.5, an open-source framework for audio-driven video generation tailored for practical applications. By focusing on empirical optimization and production readiness, we aim to narrow the gap between academic research prototypes and industrial deployments. The integration of the Whisper-large audio encoder, combined with our comprehensive data curation pipeline and scaled multi-stage training recipe, enables the model to achieve highly precise lip-synchronization, full-body temporal stability, and reliable identity consistency in long-horizon video generation.

Furthermore, our model demonstrates robust adaptability to complex real-world conditions, such as multi-person conversations, object handling, and stylized domains (e.g., anime and animals). Through the application of Group-Relative Policy Optimization (GRPO) for human preference alignment and Distribution Matching Distillation (DMD) for inference acceleration, we developed an efficient 8-NFE inference pipeline that effectively balances generation speed and visual fidelity. Extensive human evaluations across diverse scenarios indicate that LongCat-Video-Avatar 1.5 achieves highly competitive performance in naturalness, stability, and overall visual realism when compared to existing closed-source systems like OmniHuman 1.5 and HeyGen.

\paragraph{Future work.} Regarding future work, current virtual human generation models still have significant room for improvement in physical plausibility and fine-grained audio-visual synchronization. Furthermore, to maintain long-term identity consistency, existing methods often over-rely on fixed reference frames. This reliance inevitably leads to motion repetition and unnatural camera transitions constrained by the reference views. Therefore, developing a truly unbounded, infinite-length video generation framework that inherently preserves identity without rigid dependence on static reference frames remains a critical direction for future research. 

%% file: sec/7_contributors.tex
\newpage 

\section{Contributors and Acknowledgments}

All people are cataloged alphabetically by last name. (\dag) indicates the project leader and (\ddag) indicates the sponsors.

\paragraph{Contributors}\mbox{}\\[1\baselineskip]
\begin{tabular}{p{0.19\textwidth}p{0.19\textwidth}p{0.19\textwidth}p{0.19\textwidth}p{0.19\textwidth}}
Xunliang Cai\ddag & Meng Cheng & Feng Gao & Zhe Kong & Jiamu Li \\
Le Li & Weiheng Li & Hongyu Liu & Shuai Tan & Xiaoming Wei\ddag \\
Tianyu Yang & Yong Zhang\dag & & & \\
\end{tabular}

\paragraph{Acknowledgments}\mbox{}\\[1\baselineskip]
\begin{tabular}{p{0.19\textwidth}p{0.19\textwidth}p{0.19\textwidth}p{0.19\textwidth}p{0.19\textwidth}}
Fengjiao Chen & Zhuoliang Kang & Hongyu Li & Qi Li & Rumei Li \\
Shengxi Li & Shijun Liang & Xi Liu & Siyu Ren & Xuezhi Cao \\
Chao Wang & Ziwen Wang & Qilong Huang & Rixu Xie &  \\
\end{tabular}